\documentclass[runningheads]{llncs}

\usepackage{eccv}

\usepackage{eccvabbrv}

\usepackage{graphicx}
\usepackage{booktabs}
\usepackage{tabularx}
\usepackage{textcomp}
\usepackage{pifont}
\usepackage{multirow}
\usepackage{colortbl}
\newcolumntype{L}[1]{>{\raggedright\let\newline\\\arraybackslash\hspace{0pt}}m{#1}}
\newcolumntype{C}[1]{>{\centering\let\newline\\\arraybackslash\hspace{0pt}}m{#1}}
\newcolumntype{R}[1]{>{\raggedleft\let\newline\\\arraybackslash\hspace{0pt}}m{#1}}
\newcommand{\et}[2]{${#1}^{\pm{#2}}$}
\newcommand{\etb}[2]{$\mathbf{{#1}}^{\pm{#2}}$}
\newcommand{\ets}[2]{$\underline{{#1}}^{\pm{#2}}$}

\newcommand{\method}{ScaleMoGen}

\usepackage[accsupp]{axessibility}  

\usepackage[pagebackref,breaklinks,colorlinks,citecolor=eccvblue]{hyperref}

\usepackage{hyperref}

\usepackage{orcidlink}

\begin{document}

\title{\method: Autoregressive Next-Scale Prediction for Human Motion Generation}

\titlerunning{\method}

\author{Inwoo Hwang\inst{1*}\orcidlink{0009-0005-9819-1873} \and Hojun Jang\inst{1*}\orcidlink{0009-0008-0525-214X} \and 
Bing Zhou\inst{2}\orcidlink{0000-0002-0838-7858} \and Jian Wang\inst{2}\orcidlink{0000-0001-5266-3808} \and \\
Young Min Kim\inst{1\dagger}\orcidlink{0000-0002-6735-8539} \and
Chuan Guo\inst{3\dagger}\orcidlink{0000-0002-4539-0634}}

\authorrunning{I. Hwang et al.}
\institute{Seoul National University \and Snap Inc. \and Meta Reality Labs\\
\url{https://inwoohwang.me/ScaleMoGen}}

\maketitle

\def\thefootnote{*}\footnotetext{Indicates equal contribution} \def\thefootnote{$\dagger$}\footnotetext{Co-corresponding author}

\def\thefootnote{\arabic{footnote}}

\begin{abstract}
We present \method, a scale-wise autoregressive framework for text-driven human motion generation. Unlike conventional autoregressive approaches that rely on standard next-token prediction, \method~frames motion generation as a coarse-to-fine process. We quantize 3D motions into compositional discrete tokens across multiple skeletal-temporal scales of increasing granularity, learning to generate motion by autoregressively predicting next-scale token maps. To maintain structural integrity, our motion tokenizers and quantizers are explicitly designed so that discrete tokens at every scale strictly preserve the skeletal hierarchy. Additionally, we employ bitwise quantization and prediction, which efficiently scale up the tokenizer vocabulary to preserve motion details and stabilize optimization. Extensive experiments demonstrate that \method~achieves state-of-the-art performance, establishing an FID of 0.030 (vs. 0.045 for MoMask) on HumanML3D and a CLIP Score of 0.693 (vs. 0.685 for MoMask++) on the SnapMoGen dataset. Furthermore, we demonstrate that our skeletal-temporal multi-scale representation naturally facilitates training-free, text-guided motion editing.


\end{abstract}
\section{Introduction}
\label{sec:1_introduction}

Text-driven human motion generation has witnessed remarkable progress in recent years. While most existing works—such as VAEs~\cite{guo2022humanml3d, petrovich2022temos, sui2025surveyhumaninteractionmotion} and diffusion models~\cite{tevet2022mdm, chen2023mld, zhang2022motiondiffuse, zhang2023remodiffuse, meng2024mardm}—largely utilize continuous motion representations, an emerging line of work focuses on modeling motion in a discrete latent space. By quantizing motion into discrete tokens, the generation task can be reformulated as a token classification problem. This allows the model to better represent complex, multimodal motion distribution and scales effectively with large-scale datasets~\cite{van2017neural,lu2025scamo}.
Current discrete methods generally fall into two categories: autoregressive models~\cite{guo2022tm2t, zhang2023t2mgpt, jiang2023motiongpt} that generate motions through \textit{next-token} prediction, and masked models~\cite{guo2024momask, yuan2024mogents, ghosh2025duetgen, guo2025snapmogen,pinyoanuntapong2024mmm} that attend bidirectional context to iteratively fill in missing tokens within a sequence.
Recent approaches~\cite{pinyoanuntapong2024bamm,han2024bad} further combine both paradigms to leverage their complementary strengths.
In general, these approaches represent motions as temporally connected short contexts, and the central goal is to grow the token sequences along the time axis.

In this work, we propose a new perspective for generative motion modeling, shifting from \textit{temporal} expansion to \textit{next-scale} prediction, dubbed \method. This design is motivated by the inherent multi-scale structure of 3D human motions: sparse key poses already convey global semantics, while progressively denser frames and joints further add local detail~\cite{li2022ganimator,zhang2023finemogen}. This is also aligned with several multi-scale motion VQ designs, such as DuetGen~\cite{ghosh2025duetgen} and MoMask++~\cite{guo2025snapmogen}. In \method, we explicitly embed this coarse-to-fine hierarchy in both the motion representation and the generative process.

Our approach first learns a hierarchy of skeletal-temporal discrete token maps at multiple scales. A dedicated encoder maps an input motion sequence into a structured skeletal-temporal latent map, which is then progressively quantized in a residual manner into token maps of increasing skeletal-temporal granularity. Token maps across all scales are jointly supervised to reconstruct the source motion. To preserve consistent semantics across scales, we design topology-aware upsampling and downsampling operators in the VQ network and quantizers that respect the skeletal kinematic structure. At each scale, we adopt bit-wise quantization~\cite{zhao2024bsq}, representing each latent dimension as a binary label; this yields an effective vocabulary size of $2^d$ for an embedding dimension $d$.

On top of the multi-scale tokenization, a transformer autoregressively predicts the \emph{next-scale} token map conditioned on all coarser scales. Importantly, the prediction target is the binary labels rather than a single categorical index in $[0,2^d)$, which improves optimization stability. At inference time, generation starts from a $1\times1$ token map and proceeds through a constant number of scale steps, producing a full motion sequence in a coarse-to-fine manner. Beyond generation, our scale-wise skeletal-temporal model naturally supports text-driven motion editing in a zero-shot manner. Given an existing motion, users can flexibly mask tokens corresponding to specific temporal regions or skeletal parts and re-generate them conditioned on the remaining motion context and text.

Our main contributions are as follows: First, we propose \method, next-scale autoregressive framework for text-driven human motion generation. Second, we introduce a skeletal-temporal multi-scale discrete motion representation that enables structured and semantically consistent generation. Third, \method~achieves state-of-the-art motion quality and  text–motion alignment on HumanML3D\cite{guo2022humanml3d} and SnapMoGen\cite{guo2025snapmogen} benchmarks, and further supports intuitive zero-shot motion editing.
\section{Related Works}
\label{sec:2_related_works}

\subsection{Human Motion Generation}

Text-conditioned human motion generation has been actively studied for controllable animation. Early approaches relied on continuous motion representations and latent-variable generative models such as VAEs~\cite{guo2020action2motion,guo2022humanml3d, petrovich2022temos}, which enabled smooth motion synthesis but often struggled with long-horizon coherence and semantic alignment. More recently, diffusion-based models~\cite{tevet2022mdm, chen2023mld, zhang2022motiondiffuse,zhang2023remodiffuse,meng2024mardm, Bae_2025_ICCV, sfcontrol, scenemi, goaldriven, hwang2026egoforcerobustonlineegocentric} have emerged as the dominant paradigm, achieving high motion realism and strong text–motion alignment through iterative denoising. SALAD~\cite{Hong_2025_SALAD} improves generation quality by incorporating structural priors, such as skeleton-aware latent spaces, which better respect the kinematic structure of the human body.

Alongside these advances, an emerging line of work explores discrete motion representations, where continuous human motion is quantized into sequences of discrete tokens and generated using token-based generative models. Under this formulation, motion generation becomes a token classification problem, enhancing scalability and multimodal modeling~\cite{jiang2023motiongpt}.

\subsubsection{Discrete Motion Tokenization}

Most discrete motion generation methods employ vector quantization or learned codebooks to map continuous pose sequences into discrete tokens~\cite{van2017neural}. Early approaches adopt a flat, one-dimensional tokenization scheme that encodes full-body motion into temporally ordered token streams, as seen in TM2T~\cite{guo2022tm2t} and T2M-GPT~\cite{zhang2023t2mgpt}, where all joints and time steps are treated uniformly. To improve reconstruction fidelity, MoMask~\cite{guo2024momask} introduces residual vector quantization with multiple quantization layers. While effective in reducing reconstruction error, this design operates at a fixed full temporal scale, producing tokens with uneven information content and limited semantic interpretability.

More recent works move beyond flat token streams by introducing structured token layouts. MoGenTS~\cite{yuan2024mogents} organizes motion tokens on a joint–time grid to better preserve spatial correlations across body parts, while DuetGen~\cite{ghosh2025duetgen}, MoMask++~\cite{guo2025snapmogen}, and MoScale~\cite{Zheng_2026_moscale} extend earlier designs with multi-scale tokenization along the temporal dimension. Despite these advances, most existing approaches define token structures and scales in a heuristic or temporally driven manner, leaving the correspondence between token granularity and the underlying skeletal-temporal hierarchy of human motion largely implicit.

\subsection{Token-based Generative Modeling}

Given discrete motion tokens, most existing methods generate motion using either unidirectional autoregressive or masked prediction schemes. 
Unidirectional autoregressive approaches~\cite{guo2022tm2t, zhang2023t2mgpt} predict motion tokens sequentially in temporal order. This strictly sequential generation makes it difficult to capture bidirectional temporal dependencies, increases inference time, and limits overall expressiveness.
In contrast, masked prediction methods~\cite{pinyoanuntapong2024mmm, guo2024momask, yuan2024mogents, guo2025snapmogen} formulate generation as iterative token completion, but typically rely on repeated refinement and require many sampling steps at inference to achieve high-quality motion. Moreover, since generation proceeds through local token updates, these methods lack an explicit notion of global-to-local structure, making it difficult to capture global context and long-range motion dependencies.
Recent works~\cite{pinyoanuntapong2024bamm,han2024bad} attempt to unify both paradigms but still operates at a single motion scale.

Recently, advances in visual generative modeling have revisited autoregressive generation from a multi-scale perspective. Visual Autoregressive Modeling (VAR)~\cite{tian2024var, Han_2025_infinity} reformulates autoregression as coarse-to-fine next-scale prediction, replacing raster-scan next-token prediction with scale-wise generation~\cite{Zheng_2026_moscale}. This paradigm enables high-level structure to be generated first, followed by progressively finer details, improving both efficiency and structural coherence.

\subsection{Zero-Shot Motion Editing}

Learning-based motion editing has predominantly relied on paired text–motion supervision, which restricts generalization beyond the training distribution and often necessitates additional task-specific training~\cite{li2025simmotionedittextbasedhumanmotion, athanasiou2024motionfix}. To alleviate this, recent works—drawing inspiration from diffusion-based image editing techniques~\cite{meng2022sdedit, hertz2022prompt2prompt} and exploring zero-shot motion editing~\cite{Hong_2025_SALAD,kim2023flame,zhang2023finemogen} by modulating diffusion attention maps or leveraging fine-grained spatio-temporal control. However, such methods operate on a single continuous latent and do not explicitly model structural correspondence between the edited content and the original motion.

Motivated by recent next-scale autoregressive image editing approaches~\cite{Wang_2025_aredit}, our method enables zero-shot motion editing by preserving global motion context through retaining early-level motion codes, while selectively editing fine-grained details at later levels using a target text prompt. Furthermore, our approach allows precise and localized text-guided motion editing via direct replacement of motion codes associated with specific skeletal parts or temporal segments. This design highlights a key advantage of our representation in supporting structured and interpretable motion edits.
\section{\method}
\label{sec:3_method}

Our goal is to generate a 3D human pose sequence $\mathbf{m} = \{\mathbf{x}_1, \dots, \mathbf{x}_N\}$ of length $N$ guided by a textual description $c$, where each pose $\mathbf{x}_t$ is represented as a collection of joint features.
\method~encodes motions into \textit{multi-scale token maps} that explicitly reflect the \textit{skeletal-temporal hierarchy} of human movement, where coarser scales capture global semantics and finer scales model localized joint dynamics.
The model then autoregressively \textit{predicts next-scale token map} conditioned on the accumulated reconstruction from all previous coarser scales, enabling efficient coarse-to-fine generation.

\subsection{Multi-Scale Skeletal-Temporal Token Map}
\label{subsec:3_vq}

\begin{figure*}[t]
\begin{center}
\includegraphics[width=\linewidth]{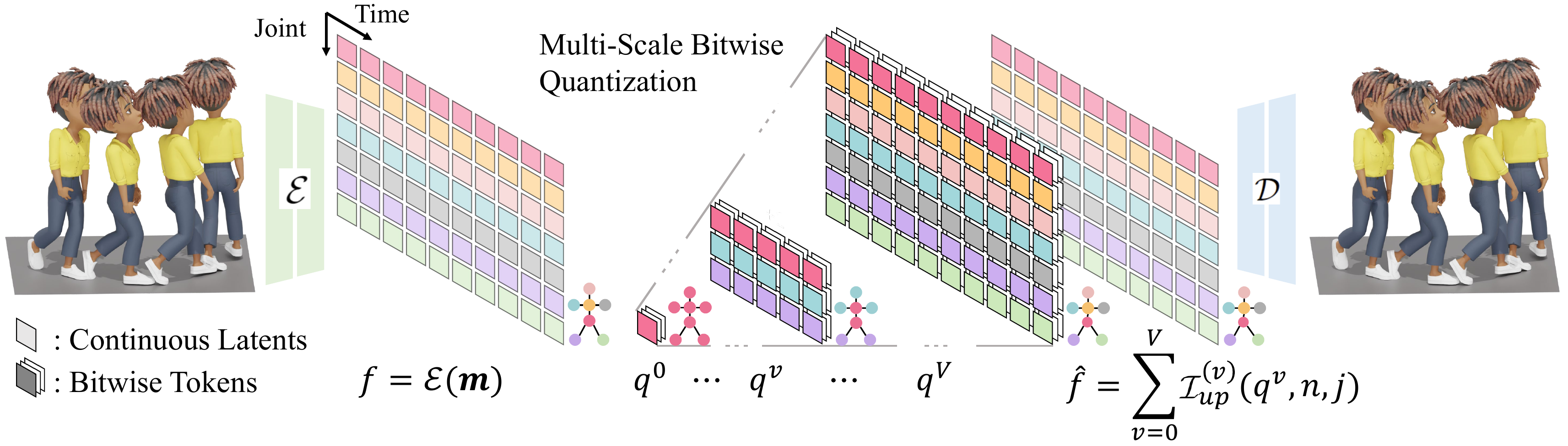}
\vspace{-2.5em}
\caption{
    Overview of our skeletal-temporal multi-scale motion quantization pipeline.
    Given an input motion sequence $\mathbf{m}$, the encoder $\mathcal{E}$ maps it to a continuous skeletal-temporal latent grid $f$.
    The latent is decomposed into a hierarchy of residual components $\{q^v\}_{v=0}^V$ via binary multi-scale residual quantization, where each scale has its own temporal resolution and skeletal partition.
    The quantized residuals are then upsampled and accumulated to form the reconstructed latent $\hat{f}$, which the decoder $\mathcal{D}$ converts back into a full-resolution motion sequence.
}
\label{fig:vq_pipeline}
\end{center}
\vspace{-2em}
\end{figure*}

Our approach first learns a structured skeletal-temporal discrete token map at multiple scales as shown in~\cref{fig:vq_pipeline}.
We first construct a skeletal-temporal feature map $f = \mathcal{E}(\mathbf{m}) \in \mathbb{R}^{n \times j \times d}$, where $n$ denotes the downsampled temporal length, $j$ the number of atomic skeletal segments, and $d$ the latent feature dimension.
We derive this latent space by extending the skeletal-temporal motion encoder $\mathcal{E}(\cdot)$ and decoder $\mathcal{D}(\cdot)$, ensuring that the latent representation captures the fundamental articulated structure of the human body.
Specifically, we maintain $j=7$ articulated atomic skeletal segments—comprising the root, spine, head, and the four extremities (both arms and both legs)—to represent this coarse skeletal structure.
Let $\mathcal{J}$ denote the set of the $j$ atomic latent segments.

\subsubsection{Skeletal-Temporal Hierarchy}

To effectively reflect the inherent multi-scale structure of human movements, we define two orthogonal axes of granularity across $(V+1)$ scales, indexed by $v \in \{0, \dots, V\}$: a sequence of temporal resolutions $h^v$ and a skeletal partition $\mathcal{S}^{(v)}$, where $|\mathcal{S}^{(v)}| = m_v$ denotes the number of semantic segments.

Temporally, we define a sequence of progressively increasing temporal resolutions $\{h^v\}_{v=0}^V$ such that $h^0 < \dots < h^V = n$.
This hierarchy ensures that the representation captures temporal motion dynamics ranging from a coarse global context at $h^0$ to a fine-grained resolution at $h^V$.

Spatially, to capture structural dependencies, we define a skeletal partition $\mathcal{S}^{(v)} = \{s_1^{(v)}, \dots, s_{m_v}^{(v)}\}$  that decomposes the atomic joint set $\mathcal{J}$ into $m_v$ semantically coherent segments.
For instance, at the coarsest scale ($v=0$), the partition represents the whole body as a single unit (i.e., $\mathcal{S}^{(0)} = \{\mathcal{J}\}$).
At intermediate scales, this decomposes into broader kinematic regions, such as the upper and lower body, to capture regional semantics.
Finally, at the finest scale ($v=V$) reaching the atomic resolution where each segment corresponds to a distinct latent body part (e.g., $\mathcal{S}^{(V)} = \{\{\texttt{\small root}\}, \{\texttt{\small spine}\}, \dots, \{\texttt{\small right leg}\}\}$).
This partitioning scheme satisfies the following properties to effectively preserve consistent semantics across scales:
\begin{itemize}
\item {Completeness:} $\bigcup_{i=1}^{m_v} s_i^{(v)} = \mathcal{J}$ and $s_a^{(v)} \cap s_b^{(v)} = \emptyset$ for $a \neq b$.
\item {Recursive Refinement:} Each segment at a finer scale is a subset of a segment at a coarser scale, i.e., $\forall s \in \mathcal{S}^{(v)}, \exists s' \in \mathcal{S}^{(v-1)}$ s.t. $s \subseteq s'$.
\end{itemize}

\subsubsection{Multi-Scale Residual Token Map}

Starting from the structured skeletal-temporal continuous feature $f \in \mathbb{R}^{n \times j \times d}$ extracted by the encoder $\mathcal{E}(\cdot)$, this latent $f$ is progressively quantized in a residual manner into $(V+1)$ multi-scale token maps $q^v \in \mathbb{R}^{h^v \times m_v \times d}$. At each level $v$, the model computes a residual feature $r^v$ in the unified global resolution and quantizes it into a token map $q^v$ with increasing skeletal-temporal granularity.

For any latent feature $z \in \mathbb{R}^{d}$, we adopt the bit-wise quantization~\cite{zhao2024bsq} operator $\mathcal{Q}_b(\cdot)$ is defined as: \begin{equation}\hat{z} = \mathcal{Q}_b(z) = \frac{1}{\sqrt{d}} \texttt{sign} \left( \frac{z}{|z|} \right)\end{equation}
This quantization represents each quantized feature as a sequence of binary codes, resulting in a scalable vocabulary size of $2^d$.

At scale level $v$, the residual feature is first downsampled to the corresponding skeletal-temporal resolution and then quantized as
\begin{equation}
    q^v = \mathcal{Q}_b\left(\mathcal{I}_{down}^{(v)}(r^v, h^v, m_v)\right), \quad r^{v+1} = r^v - \mathcal{I}_{up}^{(v)}(q^v, n, j), \quad r^0 = f
\end{equation}
where $q^v \in \mathbb{R}^{h^v \times m_v \times d}$ denotes the quantized token map at scale level $v$. Here, $\mathcal{I}_{down}^{(v)}$ and $\mathcal{I}_{up}^{(v)}$ denote the topology-aware downsampling and upsampling operators, respectively, which we describe below.
The final structured latent space $\hat{f}$ is reconstructed by accumulating these multi-scale residual contributions:
\begin{equation}
    \hat{f} = \sum_{v=0}^{V} \mathcal{I}_{up}^{(v)}(q^v, n, j) \quad \in \mathbb{R}^{n \times j \times d}
\end{equation}
This additive hierarchical structure allows \method~to successively refine motion details, starting from coarse global semantic intent to localized joint dynamics.

\subsubsection{Topology-Aware Scaling} 
Each scale token map has a different level of skeletal-temporal granularity, becoming progressively finer across scales. To accurately preserve semantics across scales while maintaining scale-specificed topology, we introduce topology-aware scaling operators, $\mathcal{I}_{down}^{(v)}$ and $\mathcal{I}_{up}^{(v)}$. This operation synchronizes semantic information between the unified global latent resolution $(n, j)$ and each topology granularity $(h^v, m_v)$ at scale $v$.

\paragraph{Topology-Aware Downsampling}($\mathcal{I}_{down}^{(v)}$): This operator maps the global residual $r^v \in \mathbb{R}^{n \times j \times d}$ to a scale-specific resolution $(h^v, m_v)$. Spatially, for each skeletal segment $s_i^{(v)} \in \mathcal{S}^{(v)}$, $i \in {1, \dots, m_v}$, it aggregates the features of its constituent joints into a single representative latent vector
Temporally, it resamples the sequence to $h^v$ temporal resolution. The resulting representation $\mathcal{I}_{down}^{(v)}(r^v, h^v, m_v)$ therefore captures the collective dynamics of each skeletal segment at the specified temporal resolution.

\paragraph{Topology-Aware Upsampling} ($\mathcal{I}_{up}^{(v)}$): This operator upscales the quantized $q^v \in \mathbb{R}^{h^v \times m_v \times d}$ back to the global unified target resolution $(n, j)$. Temporally, the sequence is upsampled to $n$ frames, while spatially each segment-level feature $s_i^{(v)}$ is broadcast to all individual joints. The resulting representation is denoted as $\mathcal{I}_{up}^{(v)}(q^v, n, j)$.

\subsubsection{Training Multi-Scale Token Map}

We optimize \method's multi-scale token map using a compound objective.
First, the reconstructed motion is obtained via the skeletal-temporal decoder $\hat{\mathbf{m}} = \mathcal{D}(\hat{f})$. The reconstruction loss measures the fidelity against the ground truth $\mathbf{m}$:
\begin{equation}
    \mathcal{L}_{\text{rec}} = \| \mathbf{m} - \hat{\mathbf{m}} \|_2^2
\end{equation}
To prevent codebook collapse and encourage diverse code usage, we adopt an entropy penalty $\mathcal{L}_{\text{ent}}$.
For a latent feature $z \in \mathbb{R}^{d}$, let $q(z)$ denote the probability distribution over binary codes induced by the quantization operator. The entropy regularization is defined as:
\begin{equation}
    \mathcal{L}_{\text{ent}} = \mathbb{E}[H(q(z))] - H[\mathbb{E}[q(z)]]
\end{equation}
where $H(\cdot)$ denotes the entropy function. The total objective is the weighted sum: $\mathcal{L}_{\text{total}} = \mathcal{L}_{\text{rec}} + \lambda \mathcal{L}_{\text{ent}}$.

\subsection{Learning Next-Scale Token Map Prediction}
\label{subsec:3_text2motion}

\subsubsection{Training}

Next, we employ a transformer $p_\phi$ to model the coarse-to-fine generation, autoregressively predicting the token map at the next scale conditioned on all coarser scales.
We utilize \texttt{T5-base}~\cite{2020t5} to transform intricate textual prompt $c$ into a sequence of word-level feature embeddings $\mathbf{c}$.
At scale $k$, the model conditions on both the text embedding $\mathbf{c}$ and a structural prefix sum $\hat{f_k}$ constructed from all preceding coarser scale token maps.
This prefix aggregates upsampled bit-tokens from earlier levels and aligns them to the current resolution through topology-aware upsampling/downsampling, providing a consistent structural context for fine-scale prediction.
\begin{equation}
    \hat{f_{k}} = \mathcal{I}_{down}^{(k)} \left( \sum_{v=0}^{k-1} \mathcal{I}_{up}^{(v)}(q^v, n, j),  h^k, m_k \right) \quad \in \mathbb{R}^{h^k \times m_k \times d}
\end{equation}
where $q^v$ represents the token map from coarse levels $v \leq k-1$, and the scaling operators $\mathcal{I}$ synchronize these residuals to the current scale’s resolution.
At the coarsest level ($k=0$), the word-level embedding $\mathbf{c}$ is projected to $\mathbb{R}^{h^0 \times m_0 \times d}$ and used as the structural prefix $\hat{f}_0$.

The transformer $p_\phi$ is trained to autoregressively predict the next-scale token map at each scale by maximizing the likelihood of the ground-truth tokens:
\begin{equation}
L_{next} = \sum_{k=0}^{V} - \log p_\phi \big(q^k \mid \hat{f}_{k}, \mathbf{c} \big).
\end{equation}
Since each token is a binary code under our quantization scheme, the likelihood decomposes into independent binary classification losses over the bits.

Our transformer backbone consists of stacked blocks for multi-scale feature aggregation.
Each block integrates self-attention, cross-attention, and feed-forward layers, together with RoPE2d~\cite{heo2024ropevit} to capture relative positional relationships over the skeletal–temporal grid.
The world-level text embeddings $\mathbf{c}$ are fused via cross-attention, guiding motion synthesis at each scale.
To mitigate error propagation across scales, we incorporate a stochastic bit-perturbation strategy during training.
With probability $r$, we randomly flip a subset of bits to simulate prediction errors, followed by re-quantization of the perturbed residual features. 
This encourages the model to recover from imperfect coarse predictions, improving robustness and enabling self-correcting behavior during generation. We further apply a block-wise causal attention mask to enforce hierarchical dependencies across scales strictly.

\subsubsection{Inference}

At inference, token maps are generated sequentially from coarse to fine scales.
For each scale $k$, the token map is sampled conditioned on the text embedding and the structural prefix made from previously formed token maps:
\begin{equation}
q^k \sim p_\phi(q^k \mid \hat{f}_k, \mathbf{c}), \quad k = 0, \dots, V.
\end{equation}

\begin{figure*}[t]
\begin{center}

\includegraphics[width=\linewidth]{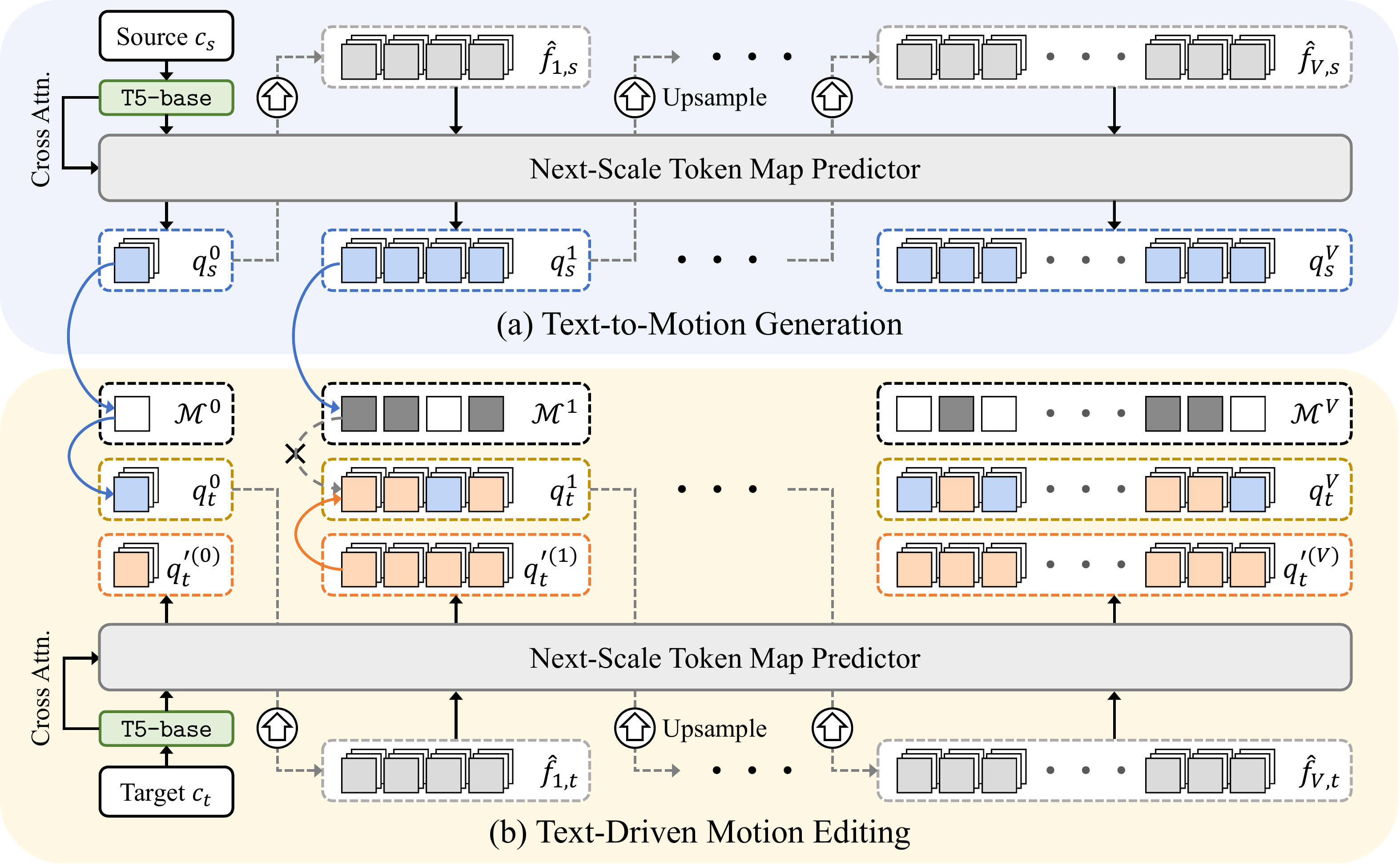}
\vspace{-2em}
\caption{
    (a) \textit{Text-to-Motion Generation:} Given a prompt $c_s$, we autoregressively predict the next-scale token maps $\{q_s^v\}_{v=0}^{V}$ conditioned on all coarser-scale token maps.
    (b) \textit{Text-Driven Motion Editing:} With additional target prompt $c_t$ with a source-token preservation mask $\{\mathcal{M}^v\}_{v=0}^{V}$, we predict edited tokens $q'^{(v)}_t$ conditioned on the remaining source motion context and $c_t$. 
    The target token maps $\{q_t^v\}_{v=0}^{V}$ are generated by blending the source tokens $q_s^v$ with the predicted tokens $q'^{(v)}_t$.
    Note that the 2D skeletal-temporal token maps are flattened into 1D for ease of visualization.
}
\label{fig:editing}
\end{center}
\vspace{-2em}
\end{figure*}

After generating all token maps, the final latent representation is reconstructed as
\begin{equation}
\hat{f} = \sum_{v=0}^{V} \mathcal{I}_{up}^{(v)}(q^v, n, j),
\end{equation}
which is then decoded to obtain the output motion sequence.
An overview of the generation process is illustrated in \cref{fig:editing} (a).

\subsection{Zero-Shot Text-Driven Motion Editing}
\label{subsec:3_editing}

By leveraging our multi-scale contextual discrete representation, which captures hierarchical skeletal structure and temporal variations, our framework supports zero-shot text-driven motion editing. 
Given a source motion $\mathbf{m}_s$, we quantize it into multi-scale discrete tokens $\{q_s^v\}_{v=0}^{V}$ and generate target tokens $\{q_t^v\}_{v=0}^{V}$ that align with a target text $c_t$ while preserving selected attributes from $\mathbf{m}_s$, which are then decoded into the edited motion $\mathbf{m}_t$.
This is achieved using a binary preservation mask $\{\mathcal{M}^v\}_{v=0}^{V} \in \{0,1\}^{h^v \times m_v}$, where $\mathcal{M}_{t,i}^{v}=1$ indicates that a token is retained from the source and $0$ indicates it should be resampled based on the $c_t$. The final mask is constructed according to three distinct criteria:

\paragraph{Global Structure Mask ($\mathcal{M}_{GS}$)}: In our multi-scale token maps, coarser scales represent global motion semantics, while finer scales capture localized details. To preserve the global structure of the original motion, we retain tokens up to a specific scale threshold $\gamma$:$$\mathcal{M}_{GS}^{v} = 1 \quad \text{if} \quad v < \gamma, \quad \text{else} \quad 0.$$ By keeping these coarse tokens, the model ensures that the edited motion $\mathbf{m}_t$ maintains the global movement trajectory of $\mathbf{m}_s$.

\paragraph{Skeletal-Temporal Editing Mask ($\mathcal{M}_{ST}$)}: A significant advantage of our method is the explicit organization of the latent space along skeletal and temporal axes. If a user intends to edit a specific set of target joints $\mathcal{J}_e \subseteq \mathcal{J}$ (e.g., the right arm) within a temporal interval $[t_0, t_1]$, the mask is defined as:
$$ \mathcal{M}_{ST}^{v}(t, i) =
\begin{cases}
0 & \text{if } s_i^{(v)} \subseteq \mathcal{J}_e \text{ and } t \in [t_0, t_1] \\
1 & \text{otherwise}
\end{cases} $$
This allows for precise, localized editing—such as modifying only the right arm while keeping the rest of the body's trajectory intact.

\paragraph{Semantic-Aware Mask ($\mathcal{M}_{SA}$)}: Inspired by~\cite{Wang_2025_aredit}, we apply a confidence-based replacement where source tokens are resampled if their likelihood under the target distribution $p(\cdot|c_t)$ is below a threshold $\tau$. This selectively updates semantically divergent regions while preserving the original motion's structural continuity.

The final preservation mask is defined as the element-wise sum of the individual masking components: $\mathcal{M}^{v} = \mathcal{M}_{GS}^{v} + \mathcal{M}_{ST}^{v} + \mathcal{M}_{SA}^{v}$. Since our model generates motion in a scale-wise manner, the target tokens $q_t^v$ at each scale $v$ are synthesized by blending the original attributes with the newly sampled distribution:
\begin{equation}
q_t^v = \mathcal{M}^{v} \odot q_s^v + (1 - \mathcal{M}^{v}) \odot q'^{(v)}_t,
\end{equation}
where $q'^{(v)}_t$ denotes token map sampled from the distribution $p(\cdot | q_{t}^{<v}, c_t)$.
An overview of the editing process is illustrated in \cref{fig:editing} (b).

\vspace{-0.5em}

\section{Experiments}
\label{sec:4_experiments}

In this section, we provide a comprehensive evaluation of \method, including text-to-motion generation (\Cref{subsec:4_t2m}), comprehensive ablation experiments (\Cref{subsec:4_ablation}), and zero-shot motion editing (\Cref{subsec:4_editing}).
Additionally, we show the sampling efficiency of \method~compared to the baselines (\Cref{subsec:4_efficiency}).
Implementation details are provided in the supplementary material.

\subsubsection{Datasets}
We evaluate \method~on two benchmarks, HumanML3D~\cite{guo2022humanml3d} and SnapMoGen~\cite{guo2025snapmogen}.
HumanML3D is a large-scale dataset built from AMASS~\cite{mahmood2019amass} and HumanAct12~\cite{guo2020action2motion}, containing 14{,}616 motion clips paired with 44{,}970 single-sentence descriptions, and serves as the standard benchmark for text-to-motion generation.
SnapMoGen is a more recent dataset comprising about 20K motion clips paired with 122K expressive and long-form descriptions that average 48 words---roughly four times longer than HumanML3D captions---making it a challenging testbed for fine-grained language conditioning.

\subsubsection{Baseline Models}
We compare \method~against a range of state-of-the-art text-to-motion methods covering three distinct training and sampling paradigms.
Diffusion-based models such as MDM~\cite{tevet2022mdm}, MLD~\cite{chen2023mld}, MotionDiffuse~\cite{zhang2022motiondiffuse}, StableMoFusion~\cite{huang2024stablemofusion}, MARDM~\cite{meng2024mardm}, and SALAD~\cite{Hong_2025_SALAD}.
Next-token prediction models in an autoregressive manner (T2M~\cite{guo2022humanml3d}, TM2T~\cite{guo2022tm2t}, T2M-GPT~\cite{zhang2023t2mgpt}), generative masked modeling (MMM~\cite{pinyoanuntapong2024mmm}, MoGenTS~\cite{yuan2024mogents}, MoMask~\cite{guo2024momask}, and MoMask++~\cite{guo2025snapmogen}), and next-scale prediction (MoScale~\cite{Zheng_2026_moscale}).

\subsubsection{Evaluation Setup}
Following standard protocols~\cite{guo2022humanml3d, guo2025snapmogen}, we report R-Precision, FID, Multimodal Distance (MM Dist), and Multimodality on HumanML3D.
For SnapMoGen, we use a TMR-style retriever~\cite{petrovich2023tmr} and report R-Precision, FID, CLIP Score, and Multimodality.
We generate one motion per text, evaluate multiple times, and report the mean with 95\% confidence intervals.
Real motions are evaluated by the same metrics and reported as an upper bound.

\subsection{Text-to-Motion Generation}
\label{subsec:4_t2m}
\begin{figure*}[t]
\begin{center}
\includegraphics[width=0.9\linewidth]{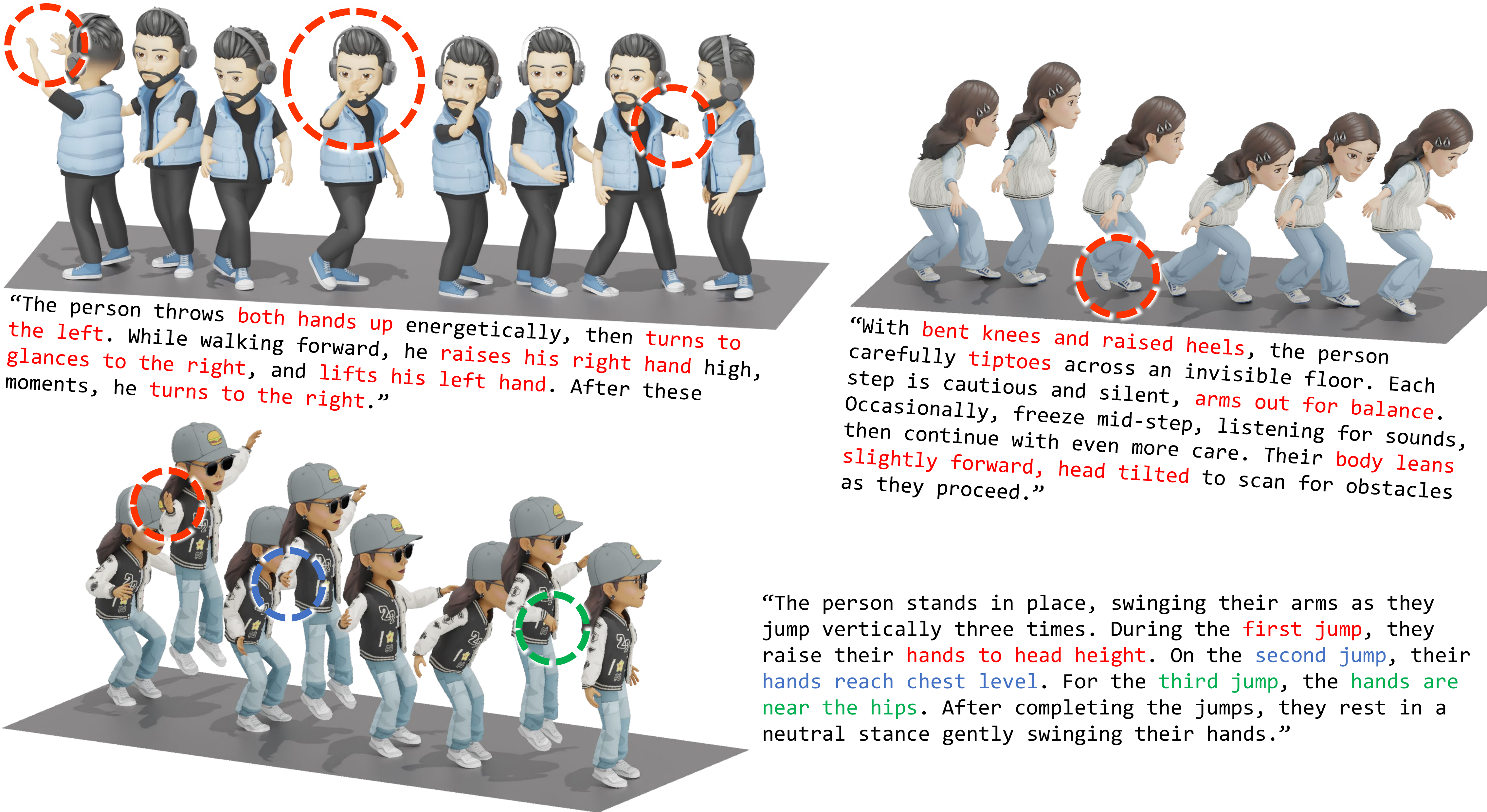}
\vspace{-0.5em}
\caption{
    Qualitative text-to-motion generation results of \method.
    Given highly descriptive, long-form text prompts, \method~accurately synthesizes complex sequences of actions (top-left), fine-grained body-part articulations (top-right), and timely executed motions with precise spatial constraints (bottom).
}
\label{fig:generation}
\end{center}
\vspace{-2.5em}
\end{figure*}
\begin{table*}[t]
    \caption{Quantitative results on  HumanML3D test set. $\pm$ indicates a 95\% confidence interval. \textbf{Bold} and \underline{underline} denote the best and second-best results, respectively.}
    \vspace{-1em}
    \centering
    % \scalebox{0.87}{
    \scalebox{0.9}{
    \begin{tabular}{l c c c c c c}
    \toprule
     \multirow{2}{*}{Methods}  & \multicolumn{3}{c}{R Precision$\uparrow$} & \multirow{2}{*}{FID$\downarrow$} & \multirow{2}{*}{MM Dist$\downarrow$} & \multirow{2}{*}{MModality$\uparrow$}\\

    \cline{2-4}
       ~ & Top 1 & Top 2 & Top 3 \\
    \midrule
    TM2T~\cite{guo2022tm2t} & \et{0.424}{.003} & \et{0.618}{.003} & \et{0.729}{.002} & \et{1.501}{.017} & \et{3.467}{.011} & \ets{2.424}{.093}  \\ 
    T2M~\cite{guo2022humanml3d} & \et{0.455}{.003} & \et{0.636}{.003} & \et{0.736}{.002} & \et{1.087}{.021} & \et{3.347}{.008} & \et{2.219}{.074}  \\   
    MDM~\cite{tevet2022mdm} & - & - & \et{0.611}{.007} & \et{0.544}{.044} & \et{5.566}{.027} & \etb{2.799}{.072}  \\
    MLD~\cite{chen2023mld} & \et{0.481}{.003} & \et{0.673}{.003} & \et{0.772}{.002} & \et{0.473}{.013} & \et{3.196}{.010} & \et{2.413}{.079}  \\
    MotionDiffuse~\cite{zhang2022motiondiffuse} & \et{0.491}{.001} & \et{0.681}{.001} & \et{0.782}{.001} & \et{0.630}{.001} & \et{3.113}{.001}  & \et{1.553}{.042}  \\
    T2M-GPT~\cite{zhang2023t2mgpt} & \et{0.492}{.003} & \et{0.679}{.002} & \et{0.775}{.002} & \et{0.141}{.005} & \et{3.121}{.009}  & \et{1.831}{.048}  \\   
    MMM~\cite{pinyoanuntapong2024mmm} & \et{0.515}{.002} & \et{0.708}{.002} & \et{0.804}{.002} & \et{0.089}{.005} & \et{2.926}{.007} & \et{1.226}{.040}  \\
    MoMask~\cite{guo2024momask} & \et{0.521}{.002} & \et{0.713}{.002} & \et{0.807}{.002} & \et{0.045}{.002} & \et{2.958}{.008} & \et{1.241}{.040}  \\
    MoGenTS~\cite{yuan2024mogents} & \et{0.529}{.003} & \et{0.719}{.002} & \et{0.812}{.002} & \ets{0.033}{.001} & \et{2.867}{.006} & \et{0.808}{.036} \\
    SALAD~\cite{Hong_2025_SALAD} & \etb{0.581}{.003} & \ets{0.769}{.003} & \etb{0.857}{.002} & \et{0.076}{.002} & \ets{2.649}{.009} & \et{1.751}{.062} \\
    MoMask++~\cite{guo2025snapmogen}  & \et{0.517}{.002} & \et{0.709}{.002} & \et{0.803}{.002} & \et{0.069}{.003} & \et{2.948}{.007} & \et{1.192}{.053}  \\
    MoScale~\cite{Zheng_2026_moscale} & \et{0.540}{.002} & \et{0.727}{.002} & \et{0.817}{.002} & \et{0.046}{.002} & \et{2.830}{.005}  & \et{0.873}{.044}  \\ 
    \midrule
    \method & \ets{0.577}{.004} & \etb{0.769}{.003} & \ets{0.856}{.002} & \etb{0.030}{.002} & \etb{2.626}{.006} & \et{1.105}{.057} \\
    \bottomrule
    \end{tabular}
    }
    \label{tab:quantitative_eval_humanml3d}
    %\vspace{-1em}

\end{table*}
\begin{table*}[t]
    \caption{Quantitative evaluation on SnapMoGen test set.}
    \vspace{-1em}
    %\bz{curious why the FID of BSQ16 is higher, did you retrain the embedding extractor?}\inwoo{Higher bit dimensions yield better FID since they preserve more information. However, predicting the bit tokens accurately becomes harder, which can lead to lower R-precision.}
    \centering
    \scalebox{0.9}{
    \begin{tabular}{l c c c c c c}
    \toprule
     \multirow{2}{*}{Methods}  & \multicolumn{3}{c}{R Precision$\uparrow$} & \multirow{2}{*}{FID$\downarrow$} & \multirow{2}{*}{CLIP Score$\uparrow$} & \multirow{2}{*}{MModality$\uparrow$}\\

    \cline{2-4}
     ~ & Top 1 & Top 2 & Top 3 \\
    \midrule
       Real motions & \et{0.940}{.001} & \et{0.976}{.001} & \et{0.985}{.001} & \et{0.001}{.000} & \et{0.837}{.000} & - \\
    \midrule
     MDM~\cite{tevet2022mdm} & \et{0.503}{.002} & \et{0.653}{.002} & \et{0.727}{.002} & \et{57.783}{.092} & \et{0.481}{.001} & \etb{13.412}{.231} \\

     T2M-GPT~\cite{zhang2023t2mgpt} & \et{0.618}{.002} & \et{0.773}{.002} & \et{0.812}{.002} & \et{32.629}{.087} & \et{0.573}{.001} & \et{9.172}{.181} \\

     StableMoFusion~\cite{huang2024stablemofusion} & \et{0.679}{.002} & \et{0.823}{.002} & \et{0.888}{.002} & \et{27.801}{.063} & \et{0.605}{.001} & \et{9.064}{.138}  \\
     MARDM~\cite{meng2024mardm} & \et{0.659}{.002} & \et{0.812}{.002} & \et{0.860}{.002} & \et{26.878}{.131} & \et{0.602}{.001} & \ets{9.812}{.287}  \\
     MoMask~\cite{guo2024momask} & \et{0.777}{.002} & \et{0.888}{.002} & \et{0.927}{.002} & \et{17.404}{.051} & \et{0.664}{.001} & \et{8.183}{.184}  \\
     MoGenTS~\cite{yuan2024mogents} & \et{0.626}{.003} & \et{0.771}{.003} & \et{0.845}{.002} & \et{27.802}{.102} & \et{0.586}{.002} & \et{7.142}{.191} \\
     SALAD~\cite{Hong_2025_SALAD} & \et{0.742}{.002} & \et{0.875}{.003} & \et{0.922}{.002} & \et{23.248}{.091} & \et{0.658}{.002} & \et{9.238}{.175} \\
     MoMask++~\cite{guo2025snapmogen} & \ets{0.802}{.001} & \ets{0.905}{.002} & \ets{0.938}{.001} & \etb{15.06}{.065} & \ets{0.685}{.001}  & \et{7.259}{.180}  \\
    \midrule
    %\method~(BSQ32) & \ets{0.795}{.006} & \ets{0.898}{.003} & \ets{0.933}{.003} & \ets{15.979}{.143} & \ets{0.683}{.001} & \ets{10.067}{.727} \\
    %\method~(BSQ16) & \etb{0.830}{.006} & \etb{0.924}{.003} & \etb{0.951}{.003} & \et{19.528}{.143} & \etb{0.705}{.001} & \et{-}{.} \\
    \method & \etb{0.807}{.004} & \etb{0.908}{.003} & \etb{0.941}{.003} & \ets{16.35}{.086} & \etb{0.693}{.001} & \et{9.399}{.669} \\
    \bottomrule
    \end{tabular}
    }
    \label{tab:quantitative_eval_snapmotion}
    \vspace{-2em}

\end{table*}

\Cref{tab:quantitative_eval_humanml3d,tab:quantitative_eval_snapmotion} show the results on HumanML3D and SnapMoGen.
On HumanML3D, \method~achieves the best FID and MM Distance. While performing comparably to SALAD~\cite{Hong_2025_SALAD} in R Precision, \method~achieves a substantially lower FID, demonstrating both high fidelity and strong text alignment.
On SnapMoGen, \method~attains the best R Precision and CLIP Score, outperforming MoMask++~\cite{guo2025snapmogen} in alignment. 
Although \method's FID is marginally higher than MoMask++, the improved CLIP Score indicates better adherence to fine-grained descriptions, confirming its generalization to rich, long-form prompts.

\Cref{fig:generation} visualizes qualitative results on complex prompts. Trained on SnapMoGen, \method~excels at interpreting fine-grained skeletal-temporal conditions.
For instance, in the top-left example, it seamlessly transitions across distinct sub-actions (throwing hands up, turning, walking).
The top-right shows nuance in body-part constraints, capturing a tiptoeing posture with bent knees. The bottom example highlights temporal precision by strictly following chronological instructions to progressively lower hand height across consecutive jumps.
These visuals align with our strong quantitative alignment scores. Please refer to our project-page for additional examples with videos.

\vspace{-1.0em}

\subsection{Component Analysis}
\label{subsec:4_ablation}

\begin{table*}[t]
    \caption{Ablation analysis of model configuration on SnapMoGen test set.}
    \vspace{-1em}
    \centering
    \scalebox{0.85}{
    \begin{tabular}{l c c c c c c c c}
    \toprule
       ~ & \multicolumn{2}{c}{VQ Config.}  & & T2M Config. & & \multicolumn{3}{c}{T2M Generation} \\
    
    \cmidrule(){2-3}
    \cmidrule(){5-5}
    \cmidrule(){7-9}
     ~ & \multirow{2}{*}{\shortstack{\footnotesize Skeleton\\-Aware}}  & \multirow{2}{*}{\footnotesize Code Size} & &  \multirow{2}{*}{\footnotesize \# params.} & & \multirow{2}{*}{\shortstack{\footnotesize R Precision$\uparrow$ \\ (Top 3)}} & \multirow{2}{*}{\footnotesize FID$\downarrow$} & \multirow{2}{*}{\footnotesize CLIP Score$\uparrow$} \\
     \\
    \midrule
    Base & \checkmark & \footnotesize 24-bit & & \footnotesize 125M & & 0.941 & 16.35 & 0.693 \\
    \midrule
    \multirow{2}{*}{\shortstack{(A) w/o Skeletal \\ Topology}} & \multirow{2}{*}{\ding{55}} & & & & & \multirow{2}{*}{0.902} & \multirow{2}{*}{26.74} & \multirow{2}{*}{0.630} \\
    \\
    \midrule

    \multirow{2}{*}{(B) Code Size} & & \footnotesize 16-bit & & & & 0.951 & 19.53& 0.705\\
    & & \footnotesize 32-bit & & & & 0.933 & 15.98& 0.683 \\
    \midrule

    % layer 16
    \multirow{3}{*}{(C) \footnotesize Larger model} & & & & \footnotesize 362M & & 0.934 & 15.84 & 0.688 \\

    % layer 24
    ~ & &  & & \footnotesize 941M & & 0.930 & 16.54 & 0.686 \\

    % layer 32
    ~ & &  & & \footnotesize 2.2B & & 0.927 & 16.51 & 0.682 \\
    \bottomrule
    \end{tabular}
    }
    \label{tab:ablation_t2m}
    \vspace{-1em}
\end{table*}

\subsubsection{Effect of Skeletal Topology}
To validate the importance of structured multi-scale quantization, we ablate the topology constraint by treating motion simply as a monolithic vector sequence rather than a structured hierarchy.
As shown in~\cref{tab:ablation_t2m} (A), \method~without skeletal topology suffers a severe performance drop across all metrics.
This confirms that embedding the inherent human skeletal structure into the latent space is crucial.
It not only provides a stronger inductive bias for realistic articulations but also creates a more expressive discrete representation that easily aligns with fine-grained descriptions.

\subsubsection{Effect of Code Size}
\Cref{tab:ablation_t2m} (B) shows that reducing code size from 24 to 16 bits improves CLIP score but degrades FID, whereas increasing to 32 bits yields the opposite trend.
A smaller code dimension reduces the effective code space, aiding the autoregressive predictor in selecting consistent text-conditioned codes (improving CLIP score).
However, the limited bit capacity increases quantization error, losing fine motion details (worsening FID).
Conversely, higher bit capacity eliminates quantization-induced degradation for better FID, but expands the discrete search space, making it harder to predict text-consistent codes.
Our base setting (24-bit) provides a good compromise between these two trade-offs.

\subsubsection{Effect of Token Map Predictor Model Size}
We investigate scaling the token map predictor from 125M to 2.2B parameters as in~\cref{tab:ablation_t2m} (C).
While motion quality lacks a consistent trend, text-motion alignment degrades with increased capacity.
This suggests that blindly scaling the autoregressive model without correspondingly expanding the dataset size leads to overfitting on the training distribution. 
The over-parameterized model memorizes specific motion sequences rather than learning generalizable text-motion mappings.
Unlike LLMs trained on Internet-scale data, current 3D motion datasets have limited volume and diversity.
Thus, the 125M model proves more robust in our setting. 
Full realization of scaling laws~\cite{henighan2020scaling,lu2025scamo} will require proportional expansion in future data scales.

\subsubsection{VQ Design and Motion Reconstruction}
We evaluate motion reconstruction metrics (FID, MPJPE) for our vector quantization (VQ) design against several alternative motion quantization and reconstruction schemes in \cref{tab:vq_ablation}. SALAD~\cite{Hong_2025_SALAD} uses continuous skeletal latents, while MoGenTS~\cite{yuan2024mogents} employs 2D residual quantization. We further compare our multi-scale skeletal-temporal bitwise VQ to a categorical codebook and a variant without skeletal topology. Our multi-scale skeletal-temporal bitwise VQ achieves the best reconstruction performance, which directly enables higher-quality motion generation. This demonstrates that combining skeletal structure with a multi-scale bitwise design is critical for preserving motion fidelity and improving generation quality.

\begin{table*}[h!]
    \centering

    \begin{minipage}{0.32\textwidth}
        \caption{VQ ablation.}
        \vspace{-1em}
        \centering
        \scalebox{0.85}{
            \begin{tabular}{l c c}
                \toprule
                Method & FID & MPJPE \\
                \midrule
                SALAD~\cite{Hong_2025_SALAD} & 0.25 & 9.91 \\
                MoGenTS~\cite{yuan2024mogents} & 1.02 & 15.34 \\
                Categorical. & 4.83 & 12.93 \\
                w/o Skeletal. & 5.60 & 6.66 \\
                \midrule
                MoScale & \textbf{0.12} & \textbf{2.73} \\
                \bottomrule
            \end{tabular}
        }
        \label{tab:vq_ablation}
    \end{minipage}\hfill
    \begin{minipage}{0.65\textwidth}
        \caption{User study on motion editing.}
    \vspace{-1em}
    \centering
    \scalebox{0.85}{
        \begin{tabular}{l c c c c c}
        \toprule
        Methods & Preservation & \phantom{ } & Semantic Align. & \phantom{ } & Overall Qual. \\
        \midrule
        MDM~\cite{tevet2022mdm} & \et{3.75}{1.35} & & \et{2.51}{1.29} & & \et{2.73}{1.22} \\
        SALAD~\cite{Hong_2025_SALAD} & \et{3.79}{1.28} & & \et{3.10}{1.45} & & \et{3.23}{1.29} \\
        \midrule
        \method & \etb{4.41}{0.78} & & \etb{4.78}{0.63} & & \etb{4.65}{0.71} \\
        \bottomrule
        \end{tabular}
        }
        \label{tab:editing}
    \end{minipage}

    \vspace{-0em}
\end{table*}

\subsection{Zero-Shot Text-Driven Motion Editing}
\label{subsec:4_editing}

\begin{figure*}[t]
\begin{center}
\includegraphics[width=0.95\linewidth]{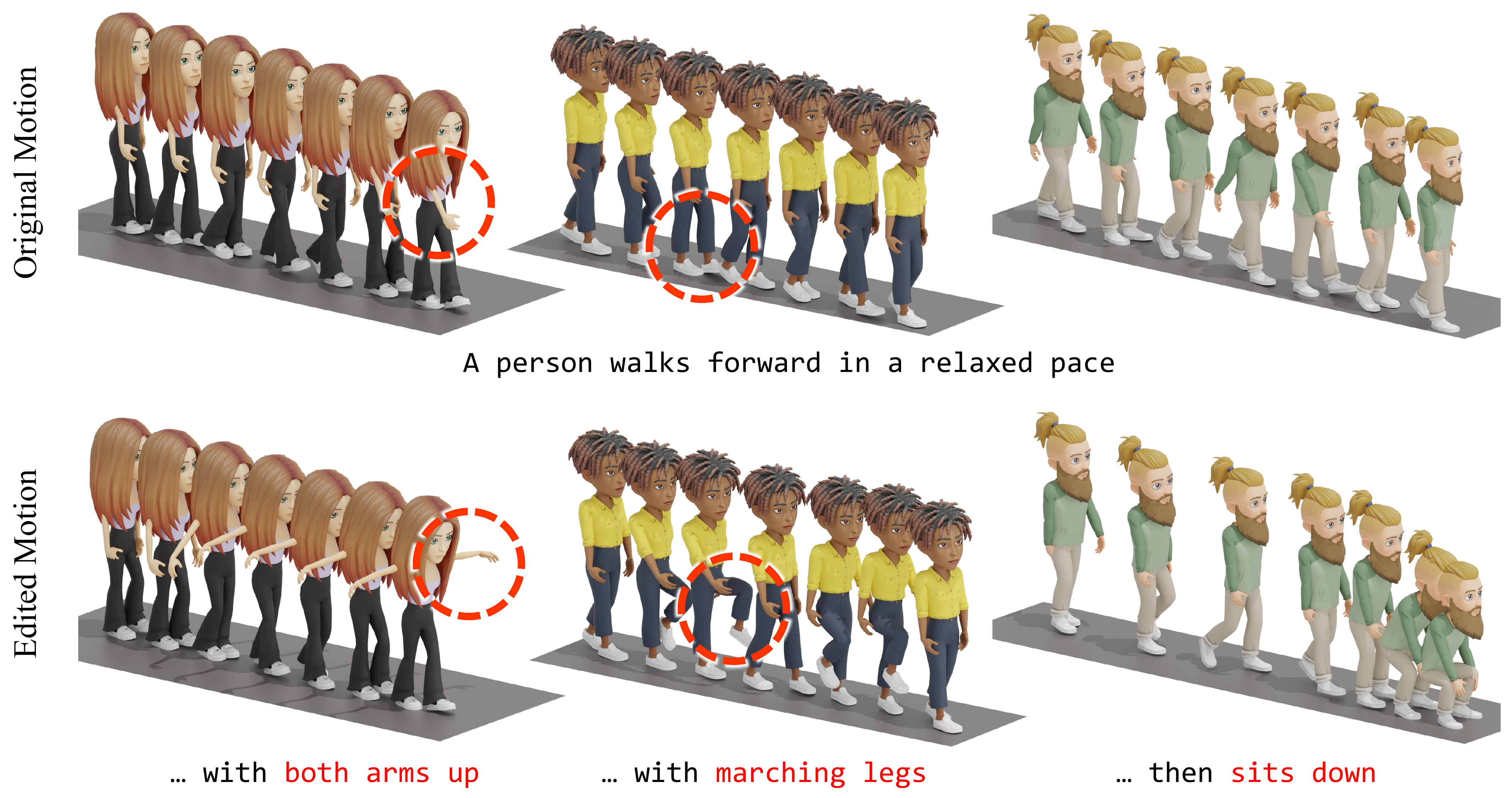}
\vspace{-0.5em}
\caption{
    Qualitative results of text-driven motion editing. 
    Given a source motion and a new target description, \method~accurately synthesizes the desired semantic changes, while preserving the identity and unrelated behaviors of the original source motion.
}
\label{fig:editing_results}
\end{center}
\vspace{-2.5em}
\end{figure*}

Since text-driven motion editing lacks established automatic metrics, we conduct a user study to evaluate editing quality.
We compare \method~against two baselines: MDM~\cite{tevet2022mdm} and SALAD~\cite{Hong_2025_SALAD}.
For each method, we first generate a source motion from the source text using its own generation pipeline, then apply the target text to produce the edited motion, yielding 10 editing examples in total.
Participants are shown three anonymized results per example and asked to rate each along three criteria on a 5-point scale: (1)~\textit{Preservation}---how well the edited motion retains aspects of the original motion unrelated to the edit; (2)~\textit{Semantic Alignment}---how accurately the edited motion reflects the target text description; and (3)~\textit{Overall Quality}---the naturalness and plausibility of the resulting motion.\\

As shown in~\cref{tab:editing}, \method~achieves the highest scores across all criteria.
Notably, our best preservation score shows that the multi-scale skeletal-temporal discretization enables precise, localized edits while leaving unrelated body parts intact.
Additionally, leading semantic alignment confirms our token replacement strategy effectively steers motion toward the target description.
These combined strengths yield the highest overall quality, proving \method~produces more natural and faithful edits than diffusion baselines.
Visual examples in \cref{fig:editing_results} further demonstrate temporal coherence and fine-grained accuracy.
Additional editing examples with videos can be found on our project-page.

\subsection{Sampling Efficiency}
\label{subsec:4_efficiency}

At inference time, the multi-scale token maps are generated autoregressively according to the predefined hierarchy with $(V+1)$ scales, requiring only $(V+1)$ sampling steps starting from the text embedding $\textbf{c}$. Table~\ref{tab:sampling_efficiency} compares the number of sampling steps and the average inference time per sample across different methods. We evaluate the computational overhead of each method on a single NVIDIA 4090 GPU. Notably, our method requires the fewest sampling steps, achieving a highly competitive runtime while maintaining superior efficiency.

\begin{table}[t]
    \caption{Sampling efficiency comparison on the SnapMoGen test set.}
    \vspace{-1em}
    \centering
    \scalebox{0.9}{
    \begin{tabular}{l c @{\hspace{1.5em}} c}
    \toprule
    Methods & Sampling Steps~$\downarrow$ & Inference Time (s)~$\downarrow$ \\
    \midrule
    MDM~\cite{tevet2022mdm} & 1000 & 10.416 \\
    T2M-GPT~\cite{zhang2023t2mgpt} & 40 & 0.239 \\
    SALAD~\cite{Hong_2025_SALAD} & 50 & 0.519 \\
    MoGenTS~\cite{yuan2024mogents} & 18 & 0.181 \\
    MoMask++~\cite{guo2025snapmogen} & \underline{10} & \textbf{0.051} \\
    \midrule
    \method & \textbf{7} & \underline{0.071} \\
    \bottomrule
    \end{tabular}
    }
    \label{tab:sampling_efficiency}
    \vspace{-1em}
\end{table}
\section{Conclusion}
\label{sec:5_conclusion}

In this work, we presented \method, a novel autoregressive next-scale prediction framework for text-driven human motion generation. 
In the \method, motion is represented as a multi-scale discrete token map and autoregressively predict next-scale token map conditioned on accumulated coarser-scale token maps. 
Specifically, our multi-scale design reflects the hierarchical structure of the human skeleton and temporal dynamics, progressively captures global motion semantics and fine-grained joint movements.
By shifting the motion generation paradigm into a scale-wise, next-scale prediction process, \method~improves text–motion alignment: high-level language governs global motion structure at coarser scales, while finer scales refine localized joint dynamics, resulting in higher overall motion quality.
Extensive experiments on HumanML3D and SnapMoGen demonstrate state-of-the-art motion fidelity and text-motion alignment.
Furthermore, the explicit correspondence between hierarchical token maps and physical motion components enables intuitive, precise zero-shot motion editing without additional training.

\section*{Acknowledgments}
This work was partially supported by the National Research Foundation of Korea (NRF) grant (No. RS-2026-25485899) and the Institute of Information \& Communications Technology Planning \& Evaluation (IITP) grant (RS-2025-25442338, AI Star Fellowship Support Program, Seoul National Univ.) funded by the Korea government (MSIT).

\bibliographystyle{splncs04}
\bibliography{main}

@String(CVPR  = {IEEE Conf. Comput. Vis. Pattern Recog.})

@String(ICCV  = {Int. Conf. Comput. Vis.})

@String(ECCV  = {Eur. Conf. Comput. Vis.})

@String(NeurIPS = {Adv. Neural Inform. Process. Syst.})

@String(AAAI  = {AAAI})

@String(ICASSP=	{ICASSP})

@String(TOG   = {ACM Trans. Graph.})

@String(CVPR  = {CVPR})

@String(ICCV  = {ICCV})

@String(ECCV  = {ECCV})

@String(NeurIPS = {NeurIPS})

@String(TOG   = {ACM TOG})

@inproceedings{guo2022tm2t,
  title={Tm2t: Stochastic and tokenized modeling for the reciprocal generation of 3d human motions and texts},
  author={Guo, Chuan and Zuo, Xinxin and Wang, Sen and Cheng, Li},
  booktitle={European Conference on Computer Vision},
  pages={580--597},
  year={2022},
  organization={Springer}
}

@inproceedings{guo2020action2motion,
  title={Action2motion: Conditioned generation of 3d human motions},
  author={Guo, Chuan and Zuo, Xinxin and Wang, Sen and Zou, Shihao and Sun, Qingyao and Deng, Annan and Gong, Minglun and Cheng, Li},
  booktitle={Proceedings of the 28th ACM International Conference on Multimedia},
  pages={2021--2029},
  year={2020}
}

@inproceedings{mahmood2019amass,
  title={AMASS: Archive of motion capture as surface shapes},
  author={Mahmood, Naureen and Ghorbani, Nima and Troje, Nikolaus F and Pons-Moll, Gerard and Black, Michael J},
  booktitle={Proceedings of the IEEE/CVF international conference on computer vision},
  pages={5442--5451},
  year={2019}
}

@article{van2017neural,
  title={Neural discrete representation learning},
  author={Van Den Oord, Aaron and Vinyals, Oriol and others},
  journal={Advances in neural information processing systems},
  volume={30},
  year={2017}
}

@article{li2022ganimator,
  title={Ganimator: Neural motion synthesis from a single sequence},
  author={Li, Peizhuo and Aberman, Kfir and Zhang, Zihan and Hanocka, Rana and Sorkine-Hornung, Olga},
  journal={ACM Transactions on Graphics (TOG)},
  volume={41},
  number={4},
  pages={1--12},
  year={2022},
  publisher={ACM New York, NY, USA}
}

@article{zhang2023finemogen,
  title={Finemogen: Fine-grained spatio-temporal motion generation and editing},
  author={Zhang, Mingyuan and Li, Huirong and Cai, Zhongang and Ren, Jiawei and Yang, Lei and Liu, Ziwei},
  journal={Advances in Neural Information Processing Systems},
  volume={36},
  pages={13981--13992},
  year={2023}
}

@inproceedings{petrovich2023tmr,
  title={Tmr: Text-to-motion retrieval using contrastive 3d human motion synthesis},
  author={Petrovich, Mathis and Black, Michael J and Varol, G{\"u}l},
  booktitle={Proceedings of the IEEE/CVF International Conference on Computer Vision},
  pages={9488--9497},
  year={2023}
}

@inproceedings{guo2022humanml3d,
  title={Generating diverse and natural 3d human motions from text},
  author={Guo, Chuan and Zou, Shihao and Zuo, Xinxin and Wang, Sen and Ji, Wei and Li, Xingyu and Cheng, Li},
  booktitle={Proceedings of the IEEE/CVF Conference on Computer Vision and Pattern Recognition},
  pages={5152--5161},
  year={2022}
}

@article{zhang2023t2mgpt,
  title={T2m-gpt: Generating human motion from textual descriptions with discrete representations},
  author={Zhang, Jianrong and Zhang, Yangsong and Cun, Xiaodong and Huang, Shaoli and Zhang, Yong and Zhao, Hongwei and Lu, Hongtao and Shen, Xi},
  journal={arXiv preprint arXiv:2301.06052},
  year={2023}
}

@inproceedings{pinyoanuntapong2024mmm,
  title={Mmm: Generative masked motion model},
  author={Pinyoanuntapong, Ekkasit and Wang, Pu and Lee, Minwoo and Chen, Chen},
  booktitle={Proceedings of the IEEE/CVF Conference on Computer Vision and Pattern Recognition},
  pages={1546--1555},
  year={2024}
}

@inproceedings{guo2024momask,
  title={Momask: Generative masked modeling of 3d human motions},
  author={Guo, Chuan and Mu, Yuxuan and Javed, Muhammad Gohar and Wang, Sen and Cheng, Li},
  booktitle={Proceedings of the IEEE/CVF Conference on Computer Vision and Pattern Recognition},
  pages={1900--1910},
  year={2024}
}

@inproceedings{petrovich2022temos,
  title={Temos: Generating diverse human motions from textual descriptions},
  author={Petrovich, Mathis and Black, Michael J and Varol, G{\"u}l},
  booktitle={European Conference on Computer Vision},
  pages={480--497},
  year={2022},
  organization={Springer}
}

@article{jiang2023motiongpt,
  title={Motiongpt: Human motion as a foreign language},
  author={Jiang, Biao and Chen, Xin and Liu, Wen and Yu, Jingyi and Yu, Gang and Chen, Tao},
  journal={Advances in Neural Information Processing Systems},
  volume={36},
  pages={20067--20079},
  year={2023}
}

@article{meng2024mardm,
      title={Rethinking Diffusion for Text-Driven Human Motion Generation},
      author={Meng, Zichong and Xie, Yiming and Peng, Xiaogang and Han, Zeyu and Jiang, Huaizu},
      journal={arXiv preprint arXiv:2411.16575},
      year={2024}
    }

@inproceedings{huang2024stablemofusion,
  title={Stablemofusion: Towards robust and efficient diffusion-based motion generation framework},
  author={Huang, Yiheng and Yang, Hui and Luo, Chuanchen and Wang, Yuxi and Xu, Shibiao and Zhang, Zhaoxiang and Zhang, Man and Peng, Junran},
  booktitle={Proceedings of the 32nd ACM International Conference on Multimedia},
  pages={224--232},
  year={2024}
}

@article{zhang2022motiondiffuse,
  title={Motiondiffuse: Text-driven human motion generation with diffusion model},
  author={Zhang, Mingyuan and Cai, Zhongang and Pan, Liang and Hong, Fangzhou and Guo, Xinying and Yang, Lei and Liu, Ziwei},
  journal={arXiv preprint arXiv:2208.15001},
  year={2022}
}

@article{tevet2022mdm,
  title={Human motion diffusion model},
  author={Tevet, Guy and Raab, Sigal and Gordon, Brian and Shafir, Yonatan and Cohen-Or, Daniel and Bermano, Amit H},
  journal={arXiv preprint arXiv:2209.14916},
  year={2022}
}

@inproceedings{chen2023mld,
  title={Executing your Commands via Motion Diffusion in Latent Space},
  author={Chen, Xin and Jiang, Biao and Liu, Wen and Huang, Zilong and Fu, Bin and Chen, Tao and Yu, Gang},
  booktitle={Proceedings of the IEEE/CVF Conference on Computer Vision and Pattern Recognition},
  pages={18000--18010},
  year={2023}
}

@article{zhang2023remodiffuse,
  title={ReMoDiffuse: Retrieval-Augmented Motion Diffusion Model},
  author={Zhang, Mingyuan and Guo, Xinying and Pan, Liang and Cai, Zhongang and Hong, Fangzhou and Li, Huirong and Yang, Lei and Liu, Ziwei},
  journal={arXiv preprint arXiv:2304.01116},
  year={2023}
}

@inproceedings{
guo2025snapmogen,
title={SnapMoGen: Human Motion Generation from Expressive Texts},
author={Chuan Guo and Inwoo Hwang and Jian Wang and Bing Zhou},
booktitle={The Thirty-ninth Annual Conference on Neural Information Processing Systems},
year={2025},
url={https://openreview.net/forum?id=pdE9onSn2h}
}

@InProceedings{Hong_2025_SALAD,
    author    = {Hong, Seokhyeon and Kim, Chaelin and Yoon, Serin and Nam, Junghyun and Cha, Sihun and Noh, Junyong},
    title     = {SALAD: Skeleton-aware Latent Diffusion for Text-driven Motion Generation and Editing},
    booktitle = {Proceedings of the Computer Vision and Pattern Recognition Conference (CVPR)},
    month     = {June},
    year      = {2025},
    pages     = {7158-7168}
}

@article{yuan2024mogents,
    title={MoGenTS: Motion Generation based on Spatial-Temporal Joint Modeling},
    author={Weihao Yuan and Weichao Shen and Yisheng HE and Yuan Dong and Xiaodong Gu and Zilong Dong and Liefeng Bo and Qixing Huang},
    journal = {Neural Information Processing Systems (NeurIPS)},
    year={2024},
}

@inproceedings{
tian2024var,
title={Visual Autoregressive Modeling: Scalable Image Generation via Next-Scale Prediction},
author={Keyu Tian and Yi Jiang and Zehuan Yuan and BINGYUE PENG and Liwei Wang},
booktitle={The Thirty-eighth Annual Conference on Neural Information Processing Systems},
year={2024},
url={https://openreview.net/forum?id=gojL67CfS8}
}

@InProceedings{Han_2025_infinity,
    author    = {Han, Jian and Liu, Jinlai and Jiang, Yi and Yan, Bin and Zhang, Yuqi and Yuan, Zehuan and Peng, Bingyue and Liu, Xiaobing},
    title     = {Infinity: Scaling Bitwise AutoRegressive Modeling for High-Resolution Image Synthesis},
    booktitle = {Proceedings of the IEEE/CVF Conference on Computer Vision and Pattern Recognition (CVPR)},
    month     = {June},
    year      = {2025},
    pages     = {15733-15744}
}

@inproceedings{
      meng2022sdedit,
      title={{SDE}dit: Guided Image Synthesis and Editing with Stochastic Differential Equations},
      author={Chenlin Meng and Yutong He and Yang Song and Jiaming Song and Jiajun Wu and Jun-Yan Zhu and Stefano Ermon},
      booktitle={International Conference on Learning Representations},
      year={2022},
}

@article{hertz2022prompt2prompt,
  title={Prompt-to-prompt image editing with cross attention control},
  author={Hertz, Amir and Mokady, Ron and Tenenbaum, Jay and Aberman, Kfir and Pritch, Yael and Cohen-Or, Daniel},
  booktitle={arXiv preprint arXiv:2208.01626},
  year={2022}
}

@InProceedings{Wang_2025_aredit,
    author    = {Wang, Yufei and Guo, Lanqing and Li, Zhihao and Huang, Jiaxing and Wang, Pichao and Wen, Bihan and Wang, Jian},
    title     = {Training-Free Text-Guided Image Editing with Visual Autoregressive Model},
    booktitle = {Proceedings of the IEEE/CVF International Conference on Computer Vision (ICCV)},
    month     = {October},
    year      = {2025},
    pages     = {17577-17586}
}

@inproceedings{kim2023flame,
  title={Flame: Free-form language-based motion synthesis \& editing},
  author={Kim, Jihoon and Kim, Jiseob and Choi, Sungjoon},
  booktitle={Proceedings of the AAAI Conference on Artificial Intelligence},
  volume={37},
  number={7},
  pages={8255--8263},
  year={2023}
}

@InProceedings{ghosh2025duetgen,
            title={DuetGen: Music Driven Two-Person Dance Generation via Hierarchical Masked Modeling},
            author={Ghosh, Anindita and Zhou, Bing and Dabral, Rishabh and Wang, Jian and Golyanik, Vladislav and Theobalt, Christian and Slusallek, Philipp and Guo, Chuan},
            booktitle={ACM SIGGRAPH},
            year={2025}
        }

@article{zhao2024bsq,
  title={Image and video tokenization with binary spherical quantization},
  author={Zhao, Yue and Xiong, Yuanjun and Kr{\"a}henb{\"u}hl, Philipp},
  journal={arXiv preprint arXiv:2406.07548},
  year={2024}
}

@inproceedings{heo2024ropevit,
    title={Rotary Position Embedding for Vision Transformer},
    author={Heo, Byeongho and Park, Song and Han, Dongyoon and Yun, Sangdoo},
    year={2024},
    booktitle={European Conference on Computer Vision (ECCV)},
}

@article{2020t5,
  author  = {Colin Raffel and Noam Shazeer and Adam Roberts and Katherine Lee and Sharan Narang and Michael Matena and Yanqi Zhou and Wei Li and Peter J. Liu},
  title   = {Exploring the Limits of Transfer Learning with a Unified Text-to-Text Transformer},
  journal = {Journal of Machine Learning Research},
  year    = {2020},
  volume  = {21},
  number  = {140},
  pages   = {1-67},
  url     = {http://jmlr.org/papers/v21/20-074.html}
}

@inproceedings{pinyoanuntapong2024bamm,
  title={BAMM: Bidirectional Autoregressive Motion Model}, 
  author={Ekkasit Pinyoanuntapong and Muhammad Usama Saleem and Pu Wang and Minwoo Lee and Srijan Das and Chen Chen}, 
  booktitle="Computer Vision -- ECCV 2024",
  year={2024},
}

@inproceedings{lu2025scamo,
  title={Scamo: Exploring the scaling law in autoregressive motion generation model},
  author={Lu, Shunlin and Wang, Jingbo and Lu, Zeyu and Chen, Ling-Hao and Dai, Wenxun and Dong, Junting and Dou, Zhiyang and Dai, Bo and Zhang, Ruimao},
  booktitle={Proceedings of the Computer Vision and Pattern Recognition Conference},
  pages={27872--27882},
  year={2025}
}

@article{li2025simmotionedittextbasedhumanmotion,
      title={SimMotionEdit: Text-Based Human Motion Editing with Motion Similarity Prediction}, 
      author={Zhengyuan Li and Kai Cheng and Anindita Ghosh and Uttaran Bhattacharya and Liangyan Gui and Aniket Bera},
      year={2025},
      eprint={2503.18211},
      archivePrefix={arXiv},
      primaryClass={cs.CV}
}

@InProceedings{scenemi,
    author    = {Hwang, Inwoo and Zhou, Bing and Kim, Young Min and Wang, Jian and Guo, Chuan},
    title     = {SceneMI: Motion In-betweening for Modeling Human-Scene Interaction},
    booktitle = {Proceedings of the IEEE/CVF International Conference on Computer Vision (ICCV)},
    month     = {October},
    year      = {2025},
    pages     = {6034-6045}
}

@misc{hwang2026egoforcerobustonlineegocentric,
      title={EgoForce: Robust Online Egocentric Motion Reconstruction via Diffusion Forcing}, 
      author={Inwoo Hwang and Donggeun Lim and Hojun Jang and Young Min Kim},
      year={2026},
      eprint={2605.13041},
      archivePrefix={arXiv},
      primaryClass={cs.CV},
      url={https://arxiv.org/abs/2605.13041}, 
}

@InProceedings{sfcontrol,
    author    = {Hwang, Inwoo and Bae, Jinseok and Lim, Donggeun and Kim, Young Min},
    title     = {Motion Synthesis with Sparse and Flexible Keyjoint Control},
    booktitle = {Proceedings of the IEEE/CVF International Conference on Computer Vision (ICCV)},
    month     = {October},
    year      = {2025},
    pages     = {13203-13213}
}

@InProceedings{Zheng_2026_moscale,
    author    = {Zheng, Zhiwei and Jin, Shibo and Liu, Lingjie and Zhao, Mingmin},
    title     = {Next-Scale Autoregressive Models for Text-to-Motion Generation},
    booktitle = {Proceedings of the IEEE/CVF Conference on Computer Vision and Pattern Recognition (CVPR)},
    month     = {June},
    year      = {2026},
    pages     = {16376-16386}
}

@InProceedings{Bae_2025_ICCV,
    author    = {Bae, Jinseok and Hwang, Inwoo and Lee, Young-Yoon and Guo, Ziyu and Liu, Joseph and Ben-Shabat, Yizhak and Kim, Young Min and Kapadia, Mubbasir},
    title     = {Less is More: Improving Motion Diffusion Models with Sparse Keyframes},
    booktitle = {Proceedings of the IEEE/CVF International Conference on Computer Vision (ICCV)},
    month     = {October},
    year      = {2025},
    pages     = {11069-11078}
}

@InProceedings{goaldriven,
  author    = {Hwang, Inwoo and Bae, Jinseok and Lim, Donggeun and Kim, Young Min},
  title     = {Goal-Driven Human Motion Synthesis in Diverse Task},
  booktitle = {Proceedings of the Computer Vision and Pattern Recognition Conference (CVPR) Workshops},
  month     = {June},
  year      = {2025},
  pages     = {2920-2930}
}

@misc{ho2022classifierfreediffusionguidance,
      title={Classifier-Free Diffusion Guidance}, 
      author={Jonathan Ho and Tim Salimans},
      year={2022},
      eprint={2207.12598},
      archivePrefix={arXiv},
      primaryClass={cs.LG},
      url={https://arxiv.org/abs/2207.12598}, 
}

@inproceedings{athanasiou2024motionfix,
  title = {{MotionFix}: Text-Driven 3D Human Motion Editing},
  author = {Athanasiou, Nikos and Ceske, Alp{\'a}r and Diomataris, Markos and Black, Michael J. and Varol, G{\"u}l},
  booktitle = {SIGGRAPH Asia 2024 Conference Papers},
  year = {2024}
}

@misc{sui2025surveyhumaninteractionmotion,
      title={A Survey on Human Interaction Motion Generation}, 
      author={Kewei Sui and Anindita Ghosh and Inwoo Hwang and Jian Wang and Chuan Guo},
      year={2025},
      eprint={2503.12763},
      archivePrefix={arXiv},
      primaryClass={cs.CV},
      url={https://arxiv.org/abs/2503.12763}, 
}

@inproceedings{han2024bad,
  title={BAD: Bidirectional Auto-regressive Diffusion for Text-to-Motion Generation},
  author={Han, S. Hossein Khatoonabadi and others},
  booktitle={IEEE International Conference on Acoustics, Speech and Signal Processing (ICASSP)},
  year={2025},
}

@article{henighan2020scaling,
  title={Scaling laws for autoregressive generative modeling},
  author={Henighan, Tom and Kaplan, Jared and Katz, Mor and Chen, Mark and Hesse, Christopher and Jackson, Jacob and Jun, Heewoo and Brown, Tom B and Dhariwal, Prafulla and Gray, Scott and others},
  journal={arXiv preprint arXiv:2010.14701},
  year={2020}
}

\clearpage

\if 0
\title{\method: Autoregressive Next-Scale Prediction for Human Motion Generation \\ \textit{Supplementary Material}}

\titlerunning{\method}

\author{Inwoo Hwang\inst{1*}\orcidlink{0009-0005-9819-1873} \and Hojun Jang\inst{1*}\orcidlink{0009-0008-0525-214X} \and 
Bing Zhou\inst{2}\orcidlink{0000-0002-0838-7858} \and Jian Wang\inst{2}\orcidlink{0000-0001-5266-3808} \and \\
Young Min Kim\inst{1\dagger}\orcidlink{0000-0002-6735-8539} \and
Chuan Guo\inst{3\dagger}\orcidlink{0000-0002-4539-0634}}

\authorrunning{I. Hwang et al.}
\institute{Seoul National University \and Snap Inc. \and Meta Reality Labs\\
\url{https://inwoohwang.me/ScaleMoGen}}

\authorrunning{I. Hwang et al.}

\maketitle
\fi

%\setcounter{equation}{10}
%\setcounter{table}{5}
%\setcounter{figure}{4}

\appendix

In the supplementary materials, we provide implementation details on the model architecture and algorithm (\Cref{sec:details}), and additional comparison results (\Cref{sec:ablations}).
We further analyze the proposed skeletal-temporal multi-scale token maps to better understand how motion information is progressively organized across scales (\Cref{sec:tokenmap_discussion}). 
For additional qualitative results, please refer to the video on our project-page.

\section{Implementation Details}
\label{sec:details}

In this section, we present the architectural and algorithmic details of our model along with the hyperparameters. The full code for training, inference, and editing will be released.

\subsection{Architectural Details}

\paragraph{Motion Encoder and Decoder}

Before quantizing human motion into a multi-scale token map, we first encode human motion using a skeletal-temporal encoder that produces a structured continuous motion feature. Our architecture follows the skeletal-temporal convolution and pooling design of SALAD~\cite{Hong_2025_SALAD} but introduces a learnable skeleton-aware pooling mechanism for more effective skeletal feature aggregation.

\begin{figure*}[t]
\begin{center}
\includegraphics[width=0.8\linewidth]{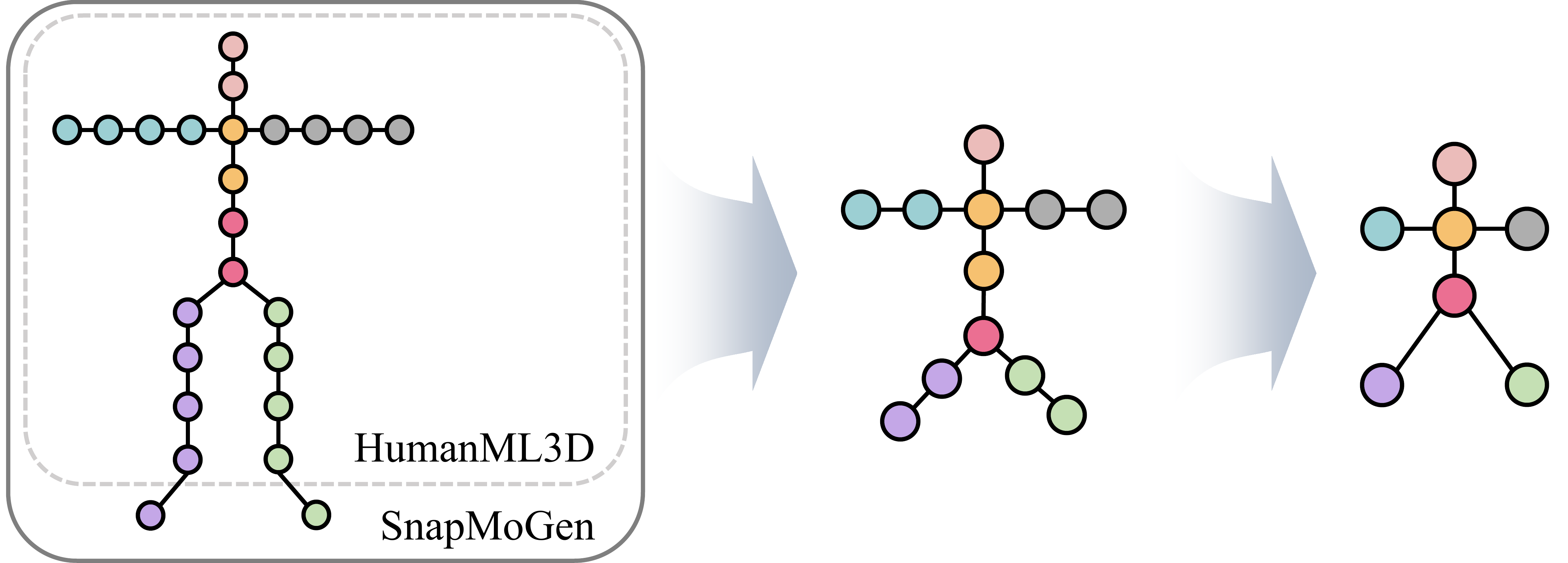}
\vspace{-0.5em}
\caption{
    The full-body skeleton is spatially downsampled by merging adjacent joints' information into coarser anatomical groups, as indicated by the color-coded regions.
    The initial full-resolution skeleton accommodates both the 22-joint HumanML3D format and the 24-joint SnapMoGen format.
    Through successive pooling stages, the model effectively captures skeletal complexity and represents the human body in an atomic 7-joint topology.
}
\label{fig:skeletal_pooling}
\end{center}
\vspace{-1em}
\end{figure*}

The encoder applies two temporal pooling stages (stride 2), resulting in a $4\times$ reduction in temporal resolution. In parallel, two learnable skeletal pooling stages aggregate joints along the kinematic hierarchy to produce the structured latent representation. This constructs a skeletal-temporal feature map $f = \mathcal{E}(\mathbf{m}) \in \mathbb{R}^{n \times j \times d}$,
where $n$ denotes the downsampled temporal length, $j$ the number of skeletal segments, and $d$ the latent feature dimension. We use $j=7$ articulated segments—root, spine, head, and four extremities—to represent a coarse skeletal structure. Details of the skeletal pooling are illustrated in~\cref{fig:skeletal_pooling}.

\paragraph{Motion Hierarchy}

We explore five skeletal-temporal hierarchy configurations (from \textcircled{1} to \textcircled{5}) to analyze the impact of skeletal structural design and skeletal-temporal scheduling. The total scale count for each strategy reflects the full hierarchy depth, combining both skeletal and temporal refinement steps.

\textcircled{1} uses a 10-scale hierarchy with a skeletal tree that initially splits the body into upper and lower halves before branching into individual limbs. It utilizes an \textit{alternating schedule}, interleaving temporal upsampling with skeletal decoupling. As the sequence's temporal resolution increases, the skeletal representation is proportionally disentangled, ensuring a balanced co-evolution of temporal frequency and localized skeletal details. 
\textcircled{2} utilizes the same lower/upper body skeletal tree but follows a \textit{temporal-first} schedule across its 10 full scales. It refines the temporal resolution to its highest fidelity while representing the entire body as a single holistic token. Only after the temporal upsampling is fully complete does the model spatially decompose the features into independent joint groups. 
\textcircled{3} employs an 11-scale hierarchy based on an anatomical ``inner-to-outer'' tree, designed to progressively decouple the core torso from the outer limbs. Using an \textit{alternating schedule}, it interleaves the progressive isolation of limb movements with step-wise temporal upsampling, allowing high-frequency limb details to emerge gradually. 
Similarly, \textcircled{4} utilizes the same 11-scale ``inner-to-outer'' skeletal tree but adopts the \textit{temporal-first} scheduling of the second strategy, increasing the temporal dimensions fully before applying skeletal decoupling at the final stages. 
Finally, \textcircled{5} uses a 7-scale full hierarchy, representing independent anatomical joint groups across all scales.

\begin{table*}[h!]
\caption{Ablation on varying motion hierarchies.
Gray-colored rows indicate the motion hierarchy configuration selected for our main experiments.
}
\vspace{-1em}
\centering
\setlength{\tabcolsep}{3pt}
\renewcommand{\arraystretch}{1.1}
\scalebox{0.8}{
    \begin{tabular}{lcccccc}
    \toprule
    \multirow{2}{*}{Motion Hierarchy}
    & \multicolumn{3}{c}{HumanML3D}
    & \multicolumn{3}{c}{SnapMoGen}\\
    \cmidrule(lr){2-4}
    \cmidrule(lr){5-7}
    & R Prec.(Top 3)$\uparrow$ & FID$\downarrow$ & MM Dist$\downarrow$
    & R Prec.(Top 3)$\uparrow$& FID$\downarrow$ & CLIP Score$\uparrow$\\
    \midrule
    \if 0
    UpLow--Alternate (10) & - & - & - & - & - & - \\
    UpLow--TempFirst (10) & - & - & - & - & - & - \\
    InOut--Alternate (11) & - & - & - & - & - & - \\
    InOut--TempFirst (11) & - & - & - & - & - & - \\
    PerJoint--TempFirst (7) & - & - & - & - & - & - \\
    \fi
    Strategy \textcircled{1} & 0.847 & 0.105 & 2.665 & 0.935 & 17.67 & 0.685 \\
    Strategy \textcircled{2} & \cellcolor{lightgray}{\textbf{0.853}} & \cellcolor{lightgray}{\textbf{0.021}} & \cellcolor{lightgray}{2.661} & 0.930 & 16.48 & 0.683 \\
    Strategy \textcircled{3} & 0.849 & 0.083 & 2.668 & 0.931 & 18.51 & 0.679 \\
    Strategy \textcircled{4} & 0.852 & 0.069 & 2.659 & 0.924 & 18.93 & 0.671 \\
    Strategy \textcircled{5} & 0.851 & 0.060 & \textbf{2.630} & \cellcolor{lightgray}{\textbf{0.941}} & \cellcolor{lightgray}{\textbf{16.35}} & \cellcolor{lightgray}{\textbf{0.693}} \\
    \bottomrule
    \end{tabular}
    }
\label{tab:hierarchy}
\end{table*}

\Cref{tab:hierarchy} reports the quantitative comparison of the five skeletal-temporal hierarchy strategies.
Due to computational setup limitations during this ablation study, the HumanML3D~\cite{guo2022humanml3d} experiments were conducted using a 32-bit latent code rather than the 24-bit dimension used in our main results, while relative performance comparisons remain valid.
Our evaluation reveals dataset-specific preferences for skeletal-temporal scheduling.
On the HumanML3D dataset, Strategy \textcircled{2} yields the best performance, improving R Precision and FID compared to the other configurations.
However, the trend differs on the SnapMoGen~\cite{guo2025snapmogen} dataset, where Strategy \textcircled{5} achieves the most favorable results across evaluation metrics.
Both results highlight that prioritizing exclusive temporal refinement over the joint disentanglement is beneficial for generating diverse and localized high-frequency motion characteristics.

\subsection{Algorithmic Details}

\paragraph{Training and Inference}

We apply a bit-perturbation strategy during training with probability $r=0.3$. In addition, a block-wise causal attention mask~\cite{tian2024var} enforces hierarchical dependencies across scales, restricting each scale token map $q^k$ to attend only to its prefixed token maps $q^0, q^1, \ldots, q^{k-1}$ and the text embedding $\mathbf{c}$.

Since motion sequences have varying lengths, we zero-pad motions to a fixed length before feeding them into the motion encoder. During bit-wise quantization, features in padded regions beyond the original motion length are forcibly quantized to $-1$. At inference time, tokens exceeding the desired length are similarly replaced with $-1$ during sampling, ensuring that the generated motion follows the specified length. All models were trained on a single NVIDIA RTX 4090 GPU.

We adopt classifier-free guidance (CFG)~\cite{ho2022classifierfreediffusionguidance} to advance text conditioning.
During training, the text embedding is randomly dropped with a probability of $10\%$ ($\mathbf{c}=\emptyset$), enabling the model to learn both conditional and unconditional predictions.
At inference time, CFG is applied at the final layer before the softmax for binary quantization.

Unlike the standard CFG that applies a constant guidance scale across all generation steps, we propose a scale-adaptive guidance strategy for our multi-scale autoregressive inference.
Specifically, the guided logits $\text{logit}_{cfg}$ at the $k$-th scale are computed as:
\begin{equation}
    \text{logit}_{cfg} = (1 + s_k) \cdot \text{logit}_{con} - s_k \cdot \text{logit}_{un}
\end{equation}
where $s_k$ is a dynamically adjusted guidance scale.
Instead of a fixed value, $s_k$ is determined by a base guidance value $S_{\text{base}}$ multiplied by the progressive ratio of the current scale: $s_k = S_{\text{base}} \times \frac{k}{V}$, where $(V+1)$ is the total number of skeletal-temporal scales, and $k \in \{0, 1, \dots, V \}$.
As shown in~\cref{tab:cfg_ablation}, the scaling progressively amplifies the conditioning signals as the model generates finer details.

\begin{table}[h]
    \caption{Ablation on classifier-free guidance}
    \vspace{-1em}
    \centering
    \scalebox{0.8}{
    \begin{tabular}{l l c c c}
    \toprule
    Methods & $S_{\text{base}}$\phantom{aaa} & \phantom{a}R Prec.(Top 3)$\uparrow$\phantom{a} & \phantom{a}FID$\downarrow$\phantom{a} & \phantom{a}CLIP Score$\uparrow$\phantom{a} \\
    
    \midrule
    w/o CFG & 0 & 0.862 & 20.86 & 0.623 \\
    Fixed CFG & 3 & 0.940 & 16.83 & 0.692 \\
    Fixed CFG & 5 & \textbf{0.943} & 17.36 & 0.693 \\
    Fixed CFG & 7 & 0.942 & 18.11 & 0.692 \\
    Decreasing CFG & 3 & 0.928 & 17.29 & 0.679 \\
    Decreasing CFG\phantom{aaa} & 5 & 0.932 & 17.22 & 0.685 \\
    Decreasing CFG & 7 & 0.932 & 17.47 & 0.686 \\
    Increasing CFG & 3 & 0.934 & 16.73 & 0.686 \\
    Increasing CFG & 7 & 0.942 & 16.93 & \textbf{0.694} \\
    \midrule
    Increasing CFG & 5 (Ours) & 0.941 & \textbf{16.35} & 0.693 \\
    \bottomrule
    \end{tabular}
    }
    \label{tab:cfg_ablation}
    \vspace{-2em}
\end{table}

\paragraph{Editing}

For the Global Structure Mask ($\mathcal{M}_{GS}$), we retain tokens up to the scale threshold $\gamma=3$ to preserve the global movement trajectory of the source motion $\mathbf{m}_s$.
For the Skeletal-Temporal Editing Mask ($\mathcal{M}_{ST}$), we manually select atomic body parts and temporal bins to keep unchanged.
For the Semantic-Aware Mask ($\mathcal{M}_{SA}$), we set the threshold $\tau=0.1$ and selectively update tokens to align the motion with the target text.

\paragraph{Topology-aware operators}

For downsampling, $\mathcal{I}_{down}^{(v)}$ maps a residual feature $r^v \in \mathbb{R}^{n \times j \times d}$ to $q^v \in \mathbb{R}^{h^v \times m_v \times d}$. We first resample along the temporal axis $(n\!\rightarrow\!h^v)$ while preserving the skeletal axis, and then average the atomic skeletal segments within each group $s_i^{(v)}$, yielding skeletal aggregation $(j\!\rightarrow\! m_v)$.

Upsampling performs the inverse operation. Each skeletal group token $q^v_{:,i} \in \mathbb{R}^{h^v \times 1 \times d},i\!=\!1,..,m_v$ is temporally upsampled to $\mathbb{R}^{n \times 1 \times d}$ and broadcast to the original atomic indices belonging to $s_i^{(v)}$. Concatenating all groups produces $\mathcal{I}_{up}^{(v)}(q^v) \in \mathbb{R}^{n \times j \times d}$.

To ensure consistent transitions across scales, operators rely strictly on group averaging. We will include these details in the revision and release the code.

\section{Additional Comparison Results}
\label{sec:ablations}

\subsubsection{Comparison with Learning-based Motion Editing}

In addition to the zero-shot editing baselines reported in paper, in \cref{tab:editing_ablation}, we also compared with direct learning-based motion editing methods using their respective dataset~\cite{athanasiou2024motionfix}, where our method consistently maintains superior performance.
Furthermore, our editing process is training-free: it does not require labeled editing data or is confined to a predefined set of editing operations, allowing flexible editing.

\begin{table}[h!]
    \centering
    \caption{User study on motion editing quality in comparison with a learning-based method.}
\scalebox{0.95}{
\begin{tabular}{lccc}
\toprule
Methods & Preservation & Semantic Alignment & Overall Quality \\
\midrule
MotionFix~\cite{athanasiou2024motionfix} & \et{3.38}{0.75} & \et{2.94}{0.75} & \et{3.16}{0.74}  \\
%\midrule
SimMotionEdit~\cite{li2025simmotionedittextbasedhumanmotion} & \et{3.71}{0.61} & \et{3.11}{0.61} & \et{3.36}{0.54}  \\
\midrule
\method  & \etb{4.29}{0.51} & \etb{4.32}{0.60}  & \etb{4.42}{0.52}  \\
\bottomrule
\end{tabular}
}
\label{tab:editing_ablation}
\vspace{-1em}
\end{table}

\subsubsection{Bit-Perturbation Probability} We additionally ablate the bit-perturbation probability $r$ on SnapMoGen~\cite{guo2025snapmogen}.
Without perturbation, the autoregressive generation process suffers from severe error accumulation. In contrast, overly strong perturbation weakens the conditioning signal. Our default setting, $r=0.3$, provides the best balance between robustness and conditional fidelity as shown in \cref{tab:bit_perturbation}.

\begin{table}[h!] 
\centering 
\caption{Ablation on the bit-perturbation probability $r$ on SnapMoGen.} 
\scalebox{0.95}{ 
\begin{tabular}{lcccccccccccc}
    \toprule 
    $r$ & \phantom{ } & 0.0 & \phantom{ } & 0.1 & \phantom{ } & 0.2 & \phantom{ } & 0.3 & \phantom{ } & 0.4 & \phantom{ } & 0.5 \\
    \midrule 
    FID $\downarrow$ & & 95.30 & & 45.64 & & 20.28 & & \textbf{16.35} & & 17.02 & & 18.27 \\ 
    R-Precision $\uparrow$ & & 0.154 & & 0.796 & & 0.903 & & \textbf{0.941} & & 0.937 & & 0.925 \\ 
    \bottomrule 
\end{tabular}
} 
\vspace{-3em}
\label{tab:bit_perturbation} \end{table}

\section{Analysis of Motion Information Across Multi-Scale Token Map}
\label{sec:tokenmap_discussion}

\begin{figure*}[h]
\begin{center}
\includegraphics[width=1.0\linewidth]{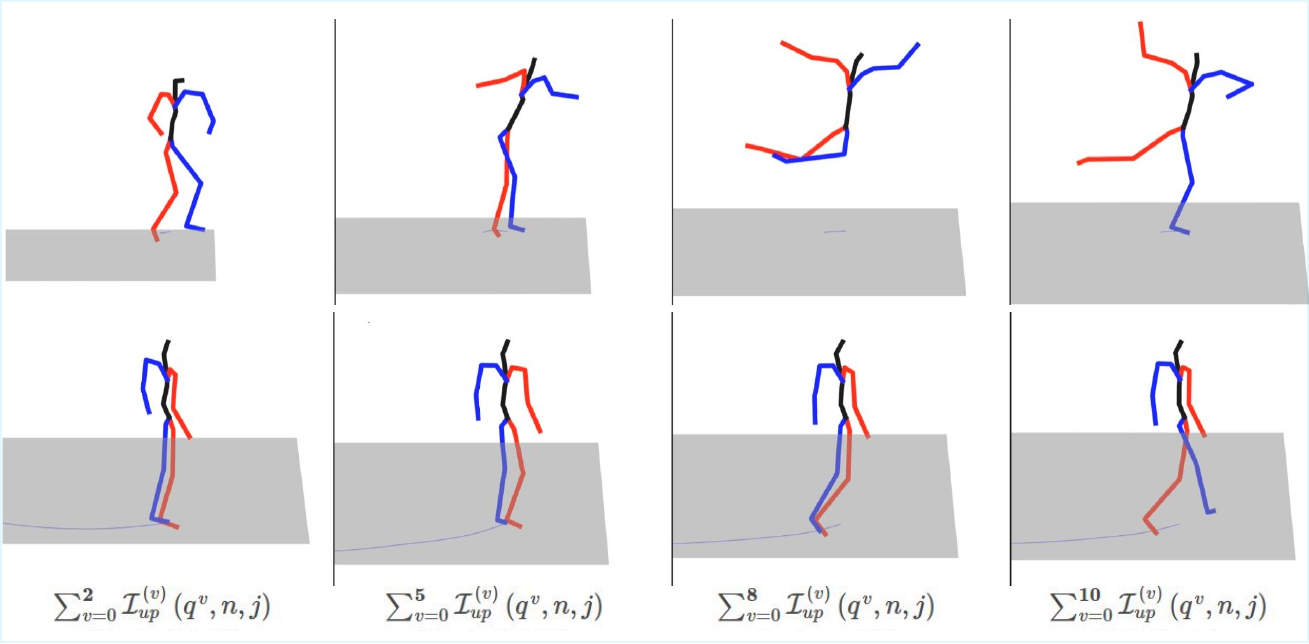}
\caption{
    Visualization of the intermediate accumulated token map motion reconstruction. Coarse token map captures global motions, which progressively disentangle into independent, fine-grained joint movements at finer scales, restoring full motion realism.
}
\label{fig:accumulated}
\end{center}
\end{figure*}

To understand how our skeletal-temporal multi-scale token map captures motion in a coarse-to-fine manner—from global motion semantics to localized joint dynamics that follow the skeletal-temporal hierarchy—we visualize intermediate representations obtained through sequential token map accumulation in \cref{fig:accumulated}. Specifically, we accumulate token maps up to scale $k$ and decode the resulting feature into the motion space.

Formally, the motion feature reconstructed from token maps up to scale $k$, and its corresponding decoded motion $\hat{\mathbf{m}}_{k}$, are given by

\begin{equation}
    \hat{f}_{k} = \sum_{v=0}^{k} \mathcal{I}_{up}^{(v)}(q^v, n, j)
    \quad \in \mathbb{R}^{n \times j \times d},
    \qquad
    \hat{\mathbf{m}}_{k} = \mathcal{D}(\hat{f}_k).
\end{equation}
where $q^v \in \mathbb{R}^{h^v \times m_v \times d}$ denotes the token map at scale $v$.

Through this visualization, we observe that coarse scales primarily capture global motion patterns such as overall body movement and trajectory. Notably, the motions of paired limbs—such as both arms or both legs—are grouped and encoded into shared representations at these coarser levels, following our predefined skeletal-temporal hierarchy.
As finer scales are progressively accumulated, more localized joint dynamics emerge, including detailed upper-body motions and leg movements. This observation confirms that the proposed skeletal-temporal hierarchy effectively organizes motion information in a coarse-to-fine manner, spanning from global motion semantics to fine-grained joint dynamics. For a more detailed visualization and understanding, please refer to our supplementary project-page for the video results.

\vspace{-3em}

%\bibliographystyle{splncs04}
%\bibliography{main}

\end{document}